\pdfoutput=1

\documentclass[11pt]{article}

\usepackage[preprint]{acl}

\usepackage{times}
\usepackage{latexsym}

\usepackage[T1]{fontenc}

\usepackage[utf8]{inputenc}

\usepackage{microtype}

\usepackage{inconsolata}

\usepackage{graphicx}

\usepackage{enumitem}
\setlist[itemize]{leftmargin=10pt, itemsep=0pt, topsep=0pt}

\title{Generating Plausible Distractors for Multiple-Choice Questions\\via Student Choice Prediction}

\author{
Yooseop Lee\textsuperscript{1,2}\thanks{This work was conducted while the first author was a graduate student at Seoul National University.}\qquad
Suin Kim\textsuperscript{2}\qquad
Yohan Jo\textsuperscript{1}\thanks{Corresponding author.}
\\  
\textsuperscript{1}Graduate School of Data Science, Seoul National University\quad
\textsuperscript{2}Elice
\\
\texttt{\{lyooseop, yohan.jo\}@snu.ac.kr}\quad
\texttt{suin@elicer.com}
}

\usepackage{tabularx} 
\usepackage{array} 
\usepackage{booktabs} 
\usepackage{ragged2e} 
\usepackage{multirow} 
\usepackage{amsmath}
\usepackage{hyperref}
\usepackage{makecell}
\usepackage{cite}
\usepackage{longtable}
\usepackage{comment}

\begin{document}
\maketitle
\begin{abstract}
In designing multiple-choice questions (MCQs) in education, creating \textit{plausible distractors} is crucial for identifying students' misconceptions and gaps in knowledge and accurately assessing their understanding.
However, prior studies on distractor generation have not paid sufficient attention to enhancing the difficulty of distractors, resulting in reduced effectiveness of MCQs.
This study presents a pipeline for training a model to generate distractors that are more likely to be selected by students. 
First, we train a \textit{pairwise ranker} to reason about students' misconceptions and assess the relative plausibility of two distractors. Using this model, we create a dataset of pairwise distractor ranks and then train a \textit{distractor generator} via Direct Preference Optimization (DPO) to generate more plausible distractors.
Experiments on computer science subjects (Python, DB, MLDL) demonstrate that our pairwise ranker effectively identifies students' potential misunderstandings and achieves ranking accuracy comparable to human experts. Furthermore, our distractor generator outperforms several baselines in generating plausible distractors and produces questions with a higher item discrimination index (DI).
\footnote{Our code and a subset of our dataset are available at \href{https://github.com/holi-lab/distractor-generator}{https://github.com/holi-lab/distractor-generator}}
\end{abstract}

\begin{figure}[t]
  \includegraphics[width=\columnwidth]{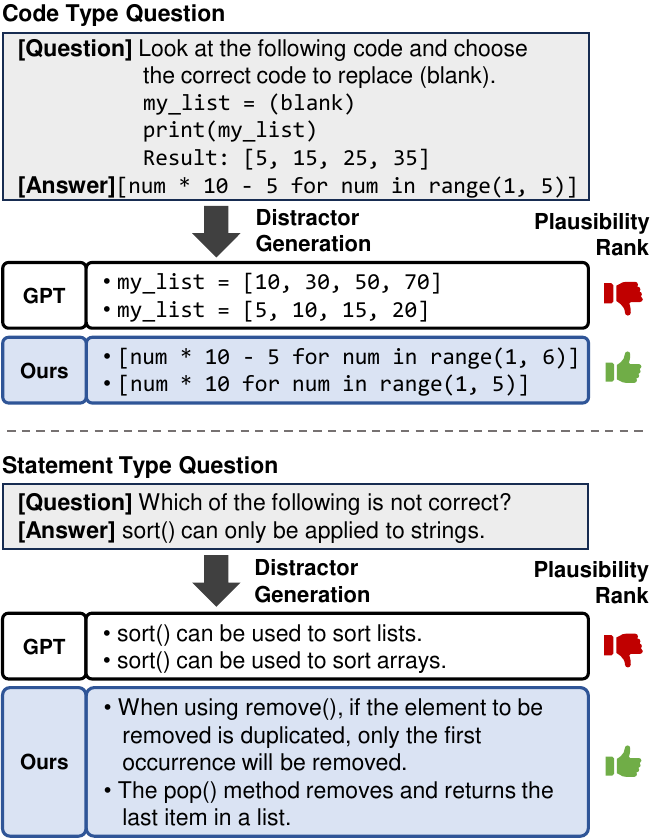}
  \caption{Examples of distractor generation. A question and a correct answer are provided as input, and the output is a set of generated distractors. The plausibility rank metric indicates how likely students are to select the distractors.}
  \label{fig:overview_distractor_generator}
\end{figure}

\section{Introduction}
Multiple-Choice Questions (MCQs) hold significant educational value as they provide a useful tool for assessing students' knowledge. 
Among the most critical elements in MCQs are \textit{distractors}---the incorrect answer options.
While the growing demand for education has amplified the need for numerous MCQs, manually creating distractors is time-consuming and costly, even for experts \citep{luo-etal-2024-chain}.
Consequently, the automation of distractor generation has emerged as a promising solution \citep{doughty-2024-comp}. 

However, prior research has focused primarily on generating distractors similar to human-authored ones \citep{fernandez-etal-2024-divert, wang-etal-2023-distractor}, with insufficient emphasis on enhancing their plausibility.
Plausible distractors are crucial as they encourage students to deliberate longer over their answers, and high-quality MCQs must possess an appropriate level of difficulty to differentiate among levels of achievement \citep{baek-2019-theory}.
By contrast, overly simplistic distractors are easily dismissed, failing to adequately assess student proficiency and reducing the educational value of the assessment.
Therefore, creating plausible distractors that target students' common mistakes or misconceptions is essential for developing highly discriminative MCQs \citep{shin-2019-multiple}.

\begin{figure*}[t]
    \centering
    \includegraphics[width=1\textwidth]{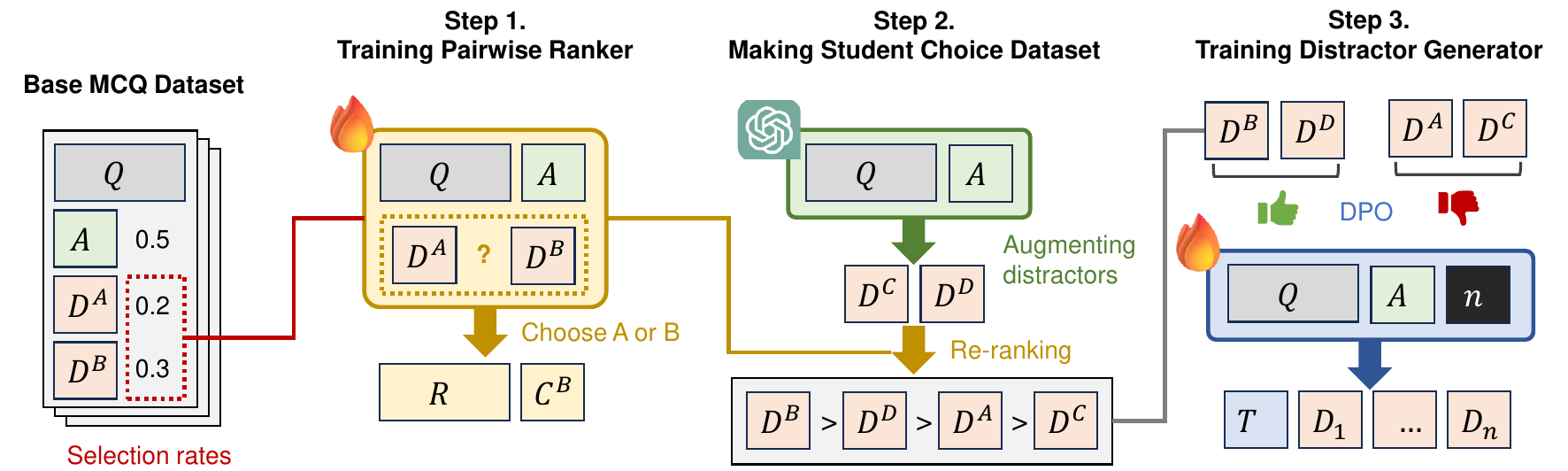}
    \caption{Training pipeline for the distractor generation.}
    \label{fig:pipeline}
\end{figure*}

Based on these needs, this study presents a model training pipeline for distractor generation. 
Figure~\ref{fig:overview_distractor_generator} illustrates example distractors generated by GPT and our model.
Our main idea is to assign relative ranks to distractors according to their likelihood of being selected by students, and use this information to train a model to generate plausible distractors. To achieve this, the process involves three steps (Figure~\ref{fig:pipeline}).
First, we train a \textit{pairwise ranker} to predict which distractors are more plausible and likely to confuse students (Step~1). 
Next, we create a synthetic \textit{student choice dataset} that includes pairwise ranking information among distractors (Step~2).
Finally, leveraging this dataset, we train a \textit{distractor generator} by applying Direct Preference Optimization (DPO, \citealp{rafailov-2024-dpo}) (Step~3).

According to evaluation on computer science (CS) subjects (Python, DB, MLDL), our pairwise ranker effectively identifies students' common misconceptions, achieving ranking accuracy comparable to human experts. In addition, the distractor generator surpasses several baselines in generating plausible distractors in both automated metrics and human studies. Notably, the distractors generated by our model exhibit a high discrimination index (DI), an essential educational metric that measures a question's ability to distinguish high-performing students from low-performing ones.

The key contributions of our study are threefold.
\begin{itemize}[itemsep=0pt, topsep=0pt]
    \item We build a pairwise ranker that reasons through students' misconceptions and predicts which distractor they are more likely to choose.
    \item We construct a student choice dataset with plausibility rankings among distractors and use it to train a plausible distractor generator.
    \item We apply our method to MCQs in CS subjects (Python, DB, MLDL) and demonstrate the generator's capability of generating distractors with high plausibility and DI.
\end{itemize}

\section{Related Works}

\subsection{Distractor Generation}
Previous studies on distractor generation can be categorized according to the question format and domain.

\paragraph{Passage-Based Format}
This format is used for exams that evaluate accurate knowledge based on provided textual content, with datasets such as RACE~\citep{lai-2017-race}, DREAM~\citep{sun-2019-dream}, SciQ~\citep{welbl-etal-2017-crowdsourcing}, and Wikipedia commonly used to generate MCQs \citep{le-berre-etal-2022-unsupervised}. As a distractor generation model for this format, \citet{offerijns-2020-better} fine-tuned GPT-2 and ensured the validity of MCQs through an external QA filtering step.
\citet{qiu-etal-2020-automatic} proposed the \textit{EDGE} framework, which reformulates passages and questions through attention mechanisms to generate distractors.
\citet{qu-etal-2024-unsupervised} introduced a dual-task training approach in which separate training was conducted using passages and questions as input to generate both answers and distractors.

However, since our study focuses on MCQs in the CS domain without relying on passages, these prior works are not directly comparable to ours.

\paragraph{Cloze-Style Format}
This format is commonly used in literacy tests and science quizzes, where test-takers fill in blanks with appropriate words \citep{chiang-etal-2022-cdgp, ren-2021-cloze}.
\citet{wang-etal-2023-distractor} proposed a \textit{pseudo Kullback-Leibler Divergence} method to regulate distractor generation by considering item discrimination factors.
\citet{yu-etal-2024-enhancing} used a knowledge graph to generate distractors by retrieving relevant triplets and selecting those most aligned with the QA context.

Our framework is not limited to cloze-style questions, which are relatively rare in our dataset, and instead supports a broader range of question types.

\paragraph{Math}
\citet{scarlatos-etal-2024-improving} improved the process of generating distractors for math problems by dividing it into two main steps: \textit{overgenerate-and-rank}. In the overgeneration phase, they used a large language model (LLM) to generate \textit{n} distractors, while in the ranking phase, a ranker was employed to select the top-\textit{k} distractors most likely to be chosen by students.
\citet{feng-2024-exploring} explored a kNN-based approach to retrieve in-context examples similar to the target question and used them to generate distractors.
\citet{fernandez-etal-2024-divert} proposed the \textit{DiVERT}, which generates distractors based on learned error representations in math MCQs. 
\citet{hangetal-2024-mcqgen} utilize retrieval-augmented generation (RAG) and chain-of-thought (CoT) for generating relevant and challenging MCQs. 

We use the methods by \citet{scarlatos-etal-2024-improving}, \citet{feng-2024-exploring} and \citet{hangetal-2024-mcqgen} as baselines for comparison with our model. We cannot compare with \citet{fernandez-etal-2024-divert} since their method requires error explanations for each distractor.

\paragraph{Other Domains}
\citet{luo-etal-2024-chain} proposed \textit{Chain-of-Exemplar Reasoning}, a method to sequentially generate distractors for multimodal questions requiring image interpretation, enhancing quality by leveraging contextually similar examples.

Meanwhile, research on distractor generation in the CS domain remains limited. While \citet{doughty-2024-comp} developed a pipeline for generating MCQs aligned with \textit{learning objectives} for programming education using GPT-4, our study emphasizes the plausibility of distractors by leveraging a smaller language model.

\subsection{Pairwise Ranker}
\label{sec:related_work_pairwise_ranker}
Our study aims to assign plausibility ranks among distractors using an LLM (Figure~\ref{fig:pipeline}, Step 1 and 2). This approach is motivated by prior findings demonstrating that LLMs exhibit strong inferential abilities, closely aligning with human performance in many evaluation tasks \citep{sun-etal-2023-chatgpt, liu-etal-2023-g}. 
Moreover, distilling these abilities from LLMs into smaller models, such as Prometheus 2 \citep{kim-2024-prometheus} fine-tuned from Mistral \citep{jiang-2023-mistral}, and ListT5 \citep{yoon-etal-2024-listt5}, has achieved comparable performance to LLMs while offering faster inference and reduced positional biases. 

However, the reasoning abilities of LLMs to rank plausible distractors in the education domain remain underexplored. A related study by \citet{scarlatos-etal-2024-improving} proposed an approach that trains a pairwise ranker for distractors using data on the actual selection rates of distractors by students. They further applied DPO to prioritize more plausible distractors. However, their model neither examines nor leverages LLMs' reasoning abilities, and the trained model lacks interpretability. In contrast, our study extensively evaluates LLMs' reasoning abilities by comparing various prompting approaches that are broadly applicable across diverse subjects. Additionally, our ranker generates reasoning behind its choices, enhancing its interpretability.

\section{Methods}
In this study, we propose a training pipeline to build a model capable of automatically generating more plausible distractors (as shown in Figure~\ref{fig:pipeline}). 
Below, we first describe the base MCQ dataset used for training (\S{\ref{sec:methods_subsection_base_mcq_dataset}}), then introduce the modeling methods for the pairwise ranker (\S{\ref{sec:methods_subsection_pairswise_ranker}}), student choice dataset (\S{\ref{sec:methods_subsection_SCD}}), and distractor generator (\S{\ref{sec:methods_subsection_distractor_generator}}).

\subsection{Base MCQ Dataset} \label{sec:methods_subsection_base_mcq_dataset}
To train both the pairwise ranker and the distractor generator, we use an MCQ dataset created by educators on a nationwide online learning platform in South Korea.
The MCQs in this dataset have been provided to K12 institutions, large corporations, and government agencies, and contain a variety of CS-related questions and student answers. We retained only those related to Python, DB (SQL), and Machine Learning \& Deep Learning (MLDL).
We target two categories of MCQs---coding and statement (see Figure~\ref{fig:overview_distractor_generator}).
The statistics of this dataset are described in Table~\ref{tab:statistics_dataset}. 

A key feature of this dataset is that it includes information on how many students answered each question and the \textit{selection rate} for each distractor. 
This allows us to determine which distractors were more confusing and plausible to students. Since each question was solved by hundreds of students from diverse sectors, the selection rate information is considered reliable.
This information will play a key role in training the pairwise ranker and distractor generator, as discussed later.
A subset of this dataset without licensing issues is available on our project website (the footnotes on the first page).

\begin{table}[t]
    \centering
    \Large
    \renewcommand{\arraystretch}{1.4}
    \resizebox{\linewidth}{!}{
    \begin{tabular}{>{\centering\arraybackslash}l|
                    >{\centering\arraybackslash}m{2.5cm}
                    >{\centering\arraybackslash}m{2.5cm}
                    >{\centering\arraybackslash}m{2.5cm}
                    >{\centering\arraybackslash}m{2.5cm}}
        \Xhline{4\arrayrulewidth}
        \shortstack{\textbf{Subject}} & 
        \shortstack{\vspace{0.2em}\\ \textbf{\# of}\\ \textbf{questions in} \\ \textbf{train/test set} \\ \vspace{0.01em}} & 
        \shortstack{\vspace{0.5em}\\ \textbf{Avg.} \\ \textbf{correctness} \\ \textbf{rate per} \\ \textbf{question} \\\vspace{0.01em}} & 
        \shortstack{\vspace{0.5em}\\ \textbf{Avg. \# of} \\ \textbf{distractors} \\ \textbf{per question} \\ \vspace{0.01em}} & 
        \shortstack{\vspace{0.5em}\\ \textbf{Avg. \# of} \\ \textbf{students} \\ \textbf{per question} \\ \vspace{0.01em}} \\ 
        \hline
        Python  & 264/52 & 70.7\% & 3.1 & 636 \\
        DB      & 54/13  & 65.6\% & 2.9 & 399 \\ 
        MLDL    & 126/32 & 61.8\% & 3.2 & 1,075 \\ 
        \Xhline{4\arrayrulewidth}
    \end{tabular}
    }
    \caption{Statistics of the base MCQ dataset. The correctness rate refers to the percentage of students who answered the question correctly.}
    \label{tab:statistics_dataset}
\end{table}

\subsection{Pairwise Ranker} \label{sec:methods_subsection_pairswise_ranker}
The pairwise ranker ($M^{Rank}$) is designed to take a question ($Q$), its correct answer ($A$), and two distractors ($D^{A}$, $D^{B}$) as input (Figure~\ref{fig:pipeline}, Step 1), and determine which distractor is more likely to be selected by students. 
\begin{equation}
  \label{eq:pairwise_ranker}
  M^{Rank}(Q, A, D^{A}, D^{B}) \rightarrow \{R, C^{A\,\text{or}\,B}\}
\end{equation}
The model outputs two main components:

\paragraph{(1) Reasoning ($R$)} 
To enhance the interpretability and accuracy of ranking results, we utilize the reasoning abilities of LLMs through a structured prompt. 
Specifically, we instruct the model to generate reasoning about (1) the knowledge being tested (e.g., ``When students approach this problem, they first need to understand ...'') based on the question and the correct answer, and (2) why each of the two given distractors might appear plausible to students (e.g., ``Distractor A might confuse students who misunderstand the syntax ...'').

\paragraph{(2) Choice ($C^{A\,\text{or}\,B}$)} 
The model outputs the result of the reasoning process as a single token (either A or B), indicating which distractor is more likely to be selected by students. 

To train a relatively small LM to perform as a ranker, we prepare some training data of reasoning for supervised fine-tuning (SFT).
Specifically, for each question in the training set of the base MCQ dataset, we prompt GPT-4o with a distractor pair and the indicator of which one was more frequently selected by students, and instruct it to generate reasoning about the two distractors that concludes in favor of the more frequently chosen one. This reasoning ($R$) and the more plausible distractor ($C^{A\,\text{or}\,B}$) form the training data for small LMs.

However, the SFT model exhibited suboptimal accuracy and became more erroneous as the reasoning grew longer.
To address this, we use DPO to further train the model’s reasoning process. 
After inference on the training set using the SFT model, samples diverging from the ground-truth choice were labeled as \textit{rejected}, while the original training samples were set as \textit{chosen}.
DPO is then applied to ensure the model generates correct reasoning and choices. 
Examples of the model's prompts are provided in Appendix~\ref{sec:appendix_PR_prompt}.

\begin{table}[t]
    \centering
    \Large
    \renewcommand{\arraystretch}{1.4}
    \resizebox{\linewidth}{!}{
    \begin{tabular}{>{\centering\arraybackslash}l|
                    >{\centering\arraybackslash}m{3.0cm}
                    >{\centering\arraybackslash}m{3.0cm}
                    >{\centering\arraybackslash}m{3.0cm}}
        \Xhline{4\arrayrulewidth}
        \shortstack{\textbf{Subject}} & 
        \shortstack{\vspace{0.5em}\\ \textbf{Avg. \# of new} \\ \textbf{distractors} \\ \textbf{in top-3} \\\vspace{0.01em}} & 
        \shortstack{\vspace{0.5em}\\ \textbf{\# of} \\ \textbf{distractor} \\ \textbf{comb. for SFT} \\\vspace{0.01em}} & 
        \shortstack{\vspace{0.5em}\\ \textbf{\# of} \\ \textbf{\textit{chosen}/\textit{rejected}} \\ \textbf{sets for DPO} \\\vspace{0.01em}} \\ 
        \hline
        All & 1.45 & 18,899 & 7,613 \\
        \Xhline{4\arrayrulewidth}
    \end{tabular}
    }
    \caption{Statistics of the student choice dataset. Columns 2 and 3 show the number of training samples  used for SFT and DPO, respectively.}
    \label{tab:statistics_SCD}
\end{table}

\subsection{Student Choice Dataset}
\label{sec:methods_subsection_SCD}
The student choice dataset is created to build training data for the distractor generator (Figure~\ref{fig:pipeline}, Step 2). 
For each question in the base MCQ dataset, GPT-4o is used to generate three new distractors distinct from the human-authored ones (Appendix~\ref{sec:appendix_SCD_augment}).
These new distractors, along with the original ones, are scored using the pairwise ranker.
At this stage, the relative rankings of the original distractors are preserved, while rankings between the original and new distractors, as well as among the new distractors, are determined by our pairwise ranker. Each question ultimately has approximately six distractors ranked in plausible order. This dataset serves for training the distractor generator for both SFT and DPO (\S{\ref{sec:methods_subsection_distractor_generator}}).

Table~\ref{tab:statistics_SCD} presents key statistics.
Column 1 of Table~\ref{tab:statistics_SCD} shows that, on average, 1.45 newly added distractors are ranked among the top 3 for each question, indicating that the newly added distractors are as plausible as the human-authored ones.

\subsection{Distractor Generator} \label{sec:methods_subsection_distractor_generator}
The distractor generator ($M^{Gen}$) takes as input a question ($Q$), its correct answer ($A$), and a hyperparameter $n$, which specifies the number of distractors to generate (Figure~\ref{fig:pipeline}, Step 3). 
The model first determines the type ($T$) of distractor (e.g., Correct/Incorrect knowledge) it will generate, and then outputs $n$ distractors ($D_{i}$).
\begin{equation}
  \label{eq:distractor generator}
  M^{Gen}(Q, A, n) \rightarrow \{T, D_{1}\, \text{...}\, D_{n}\}
\end{equation}

We ensure that the model produces distractors that are both valid and plausible as follows.

\paragraph{(1) Enhancing Validity}
Before generating distractors, the model first determines the type ($T$) of distractor. $T$ specifies whether the question requires selecting a correct or incorrect statement.
This step is critical for questions involving negation (e.g., ``Select the \textit{incorrect} statement ...'') as the model has a strong tendency to generate incorrect statements as distractors, even in such cases (see Appendix~\ref{sec:appendix_DG_additional_evaluation} for validity evaluation).

\paragraph{(2) Improving Plausibility}
To enhance the plausibility of distractors, we train the model through two stages: SFT and DPO.

\paragraph{\textbf{SFT:}} We use the student choice dataset to create training data $\{(Q, A, T, n, D_1, ..., D_n)\}$ ($n$ ranges from 1 to the maximum number of distractors available for each question). The trained model learns the basic ability to generate distractors for a given question with varying \textit{n}, but without prioritizing more plausible ones.

\paragraph{\textbf{DPO:}} To enhance the model to generate more plausible distractors, we apply DPO using the student choice dataset. 
Specifically, for each question, we construct all possible pairs between the top-$n$ distractors and the bottom-$n$ distractors, labeling the distractor from the top-$n$ as \textit{chosen} and the one from the bottom-$n$ as \textit{rejected} in each pair.
This allows the model to adjust its generation process to prioritize more plausible distractors that are more likely to challenge students.
An example of the model's prompt is provided in Appendix~\ref{sec:appendix_DG_prompt}. 
We also explored an alternative pairing method for increasing the combinations (Appendix~\ref{sec:appendix_DG_settings}), but its performance was inferior.

\section{Experiment Settings}
In this section, we describe the model training setup (\S{\ref{sec:experiments_model_training}}) and introduce the metrics used to evaluate each model (\S{\ref{sec:experiments_setup_PR}} and \S{\ref{sec:experiments_setup_DG}}).

\subsection{Model Training}
\label{sec:experiments_model_training}
For all experiments, both the pairwise ranker and the distractor generator are fine-tuned by applying LoRA \citep{hu-2021-lora} to the Mistral-7B-Instruct-v0.2. 
The numbers of training and test data are described in Table~\ref{tab:statistics_dataset} and Table~\ref{tab:statistics_SCD}. 
The detailed settings for SFT and DPO are provided in Appendix~\ref{sec:appendix_PR_settings} and \ref{sec:appendix_DG_settings}.

\subsection{Pairwise Ranker}
\label{sec:experiments_setup_PR}
\paragraph{Baselines} To assess the performance of the proposed pairwise ranker, we compare it against the following baseline models (the prompts for each baseline are included in Appendix~\ref{sec:appendix_PR_prompt}):
\begin{itemize}
    \item \textbf{GPT-3.5-turbo and GPT-4o:}
    We instruct these GPT models to predict the ranking between two distractros in a zero-shot manner.
    To examine the impact of different prompt formats, we experiment with four approaches: (1) \textbf{Reasoning}: the reasoning-based prompt format described in \S{\ref{sec:methods_subsection_pairswise_ranker}},  (2) \textbf{Rubric}: scoring based on evaluation criteria for assessing plausibility, (3) \textbf{G-Eval}: adapting the prompt proposed by \citet{liu-etal-2023-g} for our specific task, and (4) \textbf{Discussion}: simulating a collaborative learning scenario where two teacher agents discuss while observing students' problem-solving processes.
    \item \textbf{\citet{scarlatos-etal-2024-improving}:} We follow the pairwise ranker prompt and training/inference method proposed in this paper, replacing their data with ours.
\end{itemize}

\paragraph{Training Data} We use two distinct settings for training data (Table~\ref{tab:statistics_dataset}):
\begin{itemize}
    \item \textbf{Separate (Sep.):} Models trained separately with data for each subject—Python, DB, and MLDL.
    \item \textbf{Combined (Comb.):} A model trained with data from all subjects combined.
\end{itemize}

\paragraph{Distractor Order}
One known limitation of LLM-based pairwise ranking is \textit{positional bias}, where the output may vary depending on whether two choices, A and B, are presented in the input prompt as AB or BA \citep{yoon-etal-2024-listt5}.
To address this, we set the temperature to 0.5 and repeat the reasoning process with both AB and BA input sequences until consistent outputs are achieved, or randomly select a result after 10 attempts.

\paragraph{Evaluation Metrics}
The evaluation metrics for the pairwise ranker are as follows:

\begin{itemize}
\item \textbf{Rank Accuracy} measures how often the ranker correctly identifies the distractor with the higher student selection rate in the test set.

\item \textbf{Human Evaluation} aims to compare the model's performance with human experts. First, two professors in data science perform the pairwise ranking task on 60 test samples (20 per subject), and their results are compared with our model's rank accuracy.
Second, three Master's students majoring in data science assess the quality of model-generated reasoning and ranking results. For this, 30 samples (10 per subject) of reasoning and choices generated by our pairwise ranker (`DPO, Comb.' in Table~\ref{tab:evaluation_PR_rank_acc}) are randomly selected from the test set. 
The survey form and the rubric are in Appendix~\ref{sec:appendix_PR_human_evaluation}.

\item \textbf{Consistency in Rank Prediction} tracks the number of iterations required for the model to predict the same choice for both AB and BA inputs. 
Fewer iterations indicate lower positional bias.
\end{itemize}

\subsection{Distractor Generator}
\label{sec:experiments_setup_DG}
The performance of our distractor generator is evaluated using the following metrics:

\paragraph{(1) Plausibility} 
We compare the plausibility of distractors generated by our model, GPT models, a kNN approach \citep{feng-2024-exploring}, a CoT prompting approach \citep{hangetal-2024-mcqgen}, and human experts (from the base MCQ dataset) as measured by our pairwise ranker (`DPO, Comb.' in Table~\ref{tab:evaluation_PR_rank_acc}). 
Win/tie/lose counts are calculated per question/distractor in two settings:
\begin{itemize}
    \item \textbf{Setting A:} For each test question, three distractors are generated by each model ($n=3$), and only valid ones are retained. These are then compared pairwise between two models, with one point awarded to the winner. Identical distractors are excluded from comparisons. 
    \item \textbf{Setting B:} 
    To account for cases where models generate fewer than three valid distractors, each model's temperature is increased to generate up to five valid distractors per model. After excluding identical distractors between the models, the top-3 are selected for pairwise comparison.
\end{itemize}

\paragraph{(2) Human Evaluation} 
We conduct a human evaluation where actual students assess the difficulty of distractors generated by our method.
The test comprises 40 MCQs (Python: 20, DB: 10, MLDL: 10). Each question was sampled from the test set of the base MCQ dataset and paired with four distractors, one from each model (SFT, DPO, GPT-3.5-turbo, and GPT-4o), along with a `None of the above' option. 
The test is taken by 15 college students enrolled in AI courses at our university\footnote{The sample size is larger than the one tested on three individuals in \citet{luo-etal-2024-chain}.}.
Based on the selection counts for each distractor, we calculate the plausibility and discrimination index for each model. The discrimination index indicates the ability of each item to differentiate between high- and low-performing students and is calculated as $DI = (U - L) / N$, where $U$ and $L$ denote the number of students in the upper ($U$) and lower ($L$) groups who answered the item correctly, and $N$ is the number of students in each group.

We also evaluated the \textit{clarity} (i.e., whether each distractor is clearly written without ambiguity) and \textit{answerability} (i.e., whether a student with relevant knowledge can reasonably answer the question, as defined by \citet{moon-etal-2022-evaluating}) of MCQs composed solely of the distractors generated by our DPO model with 11 Master's students in data science.
More details about the human evaluation are provided in Appendix~\ref{sec:appendix_DG_human_evaluation}.

\section{Experiment Results}
\label{sec:experiments_result}
In this section, we present the experimental results for the pairwise ranker (\S{\ref{sec:experiments_result_PR}}) and the distractor generator (\S{\ref{sec:experiments_result_DG}}).

\subsection{Pairwise Ranker}
\label{sec:experiments_result_PR}

\begin{table}[t]
    \centering
    \Large
    \renewcommand{\arraystretch}{1.5}
    \setlength{\arrayrulewidth}{0.15pt}
    \resizebox{\columnwidth}{!}{
    \begin{tabular}{l|cccc}
        \Xhline{8\arrayrulewidth}
        & \multicolumn{4}{c}{\textbf{Rank Accuracy $\uparrow$}} \\ \cline{2-5}
        & Python & DB & MLDL & Avg. \\
        \hline
        GPT-3.5 (Reasoning) & 0.633 & 0.523 & 0.606 & 0.587 \\
        GPT-4o (Reasoning) & 0.686 & 0.664 & 0.570 & 0.640 \\
        GPT-4o (Reasoning, 3-shot) & 0.674 & \textbf{0.673} & 0.600 & 0.649 \\
        GPT-4o (Rubric) & 0.686 & 0.500 & 0.624 & 0.603 \\
        GPT-4o (G-Eval) & 0.632 & 0.550 & 0.543 & 0.575 \\
        GPT-4o (Discussion) & 0.549 & 0.482 & 0.487 & 0.506 \\
        \citet{scarlatos-etal-2024-improving} & 0.532 & 0.386 & 0.545 & 0.488 \\
        \hline
        Ours (SFT, Sep.) & 0.677 & 0.491 & 0.594 & 0.587 \\
        Ours (SFT, Comb.) & 0.642 & 0.650 & \textbf{0.677} & 0.657 \\
        Ours (DPO, Comb.) & \textbf{0.712} & 0.659 & 0.655 & \textbf{0.675} \\
        \hline
        Ours (SFT w/o Reasoning) & 0.659 & 0.523 & 0.521 & 0.567 \\
        \Xhline{8\arrayrulewidth}
    \end{tabular}
    }
    \setlength{\arrayrulewidth}{0.4pt}
    \caption{Evaluation results on pairwise rankers. The results were averaged over five generations for each model.}
    \label{tab:evaluation_PR_rank_acc}
\end{table}

\paragraph{(1) Rank Accuracy}
As shown in Table~\ref{tab:evaluation_PR_rank_acc}, in terms of accuracy, our DPO model achieved an accuracy of 67.5\% (row 10), outperforming GPT-3.5-turbo (58.7\%, row 1) and GPT-4o (64.0\%, row 2) on average. This result is somewhat surprising because our model was trained on reasoning generated by GPT-4o.
Moreover, the DPO model significantly outperformed the SFT models (58.7\%--65.7\%, rows 8--9), particularly in Python, showing the effectiveness of DPO in enhancing the reasoning capability of the model.
While \citet{scarlatos-etal-2024-improving}'s method achieved strong performance on math questions in their original work, it exhibited lower accuracy on the CS subjects (48.8\%, row 7).

\begin{figure}[t]
    \centering
    \includegraphics[width=1\columnwidth]{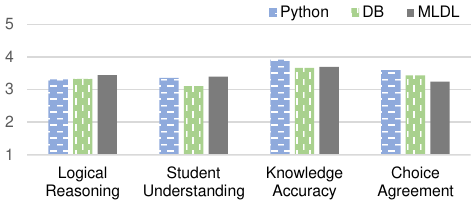}
    \caption{Human evaluation on our pairwise ranker. The results from participants were averaged.}
    \label{fig:DG_human_evaluation}
\end{figure}

\paragraph{(2) Human Evaluation}
Human experts (two professors) tasked with choosing the more plausible distractor for 60 questions achieved an accuracy of 71.7\%, compared to 70\% achieved by our DPO model on the same task. This result suggests that the task is challenging even for experts and that GPT-like LLMs trained on large data can predict the confusion experienced by students at a level comparable to human performance. 

Figure~\ref{fig:DG_human_evaluation} presents survey results from three Master's students evaluating the reasoning quality of the DPO model on a 5-point Likert scale.
These results provide mild to moderate evidence supporting the model’s ability to infer students’ misconceptions through logical reasoning and accurate knowledge.

\begin{figure}[t]
    \centering
    \includegraphics[width=1\columnwidth]{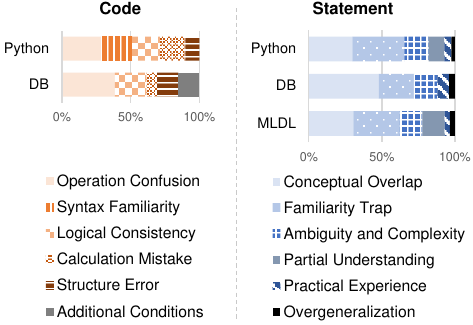}
    \caption{Plausibility factors in our pairwise ranker's reasoning.}
    \label{fig:plausibility_factors}
\end{figure}

\paragraph{(3) Plausibility Factors}
We analyzed main factors revealed in the model's reasoning to determine plausibility. We selected reasoning outputs where the DPO model predicted correct choices, and categorized plausibility factors in collaboration with GPT-4o. Figure~\ref{fig:plausibility_factors} visualizes the proportion of each category. 
In the code type questions (e.g., determining the output of a code snippet or filling in blanks), factors such as incorrect assumptions about function outputs or operations were the most common, while in the statement type questions (e.g., selecting statements about concepts), factors like conceptual overlap with other similar terms appeared most frequently.
Definitions for each category can be found in Appendix~\ref{sec:appendix_PR_factors}.

\paragraph{(4) Reasoning Methods}
We conducted an ablation study to examine the effectiveness of our reasoning method for rank accuracy. As shown in Table~\ref{tab:evaluation_PR_rank_acc}, for GPT-4o, using our reasoning structure (row 2) substantially outperformed other reasoning methods (rows 4--6), leading us to adopt the current reasoning format for the trained models. 
Training the model without the reasoning process (row 11) significantly reduced ranking accuracy, highlighting the importance of our reasoning method. 

\paragraph{(5) Consistency in Rank Prediction}
We evaluated the consistency of predictions when input order was altered and found that our model exhibits lower positional bias compared to GPT-3.5-turbo. The experimental results are provided in Appendix~\ref{sec:appendix_PR_generation_attempts}.

\paragraph{(6) Error Analysis}
Upon analyzing cases where our pairwise ranker produced incorrect reasoning, we identified several types of error, such as misjudging implausible errors as plausible and struggling with reasoning for unfamiliar questions that were underrepresented in the training data.
A detailed analysis and suggestions for future work can be found in Appendix~\ref{sec:appendix_PR_error_analysis}.

\subsection{Distractor Generator}
\label{sec:experiments_result_DG}

\begin{table}[t]
    \centering
    \Large
    \renewcommand{\arraystretch}{1.4}
    \resizebox{\columnwidth}{!}{
    \begin{tabular}{ll|cc|cc} 
        \Xhline{4\arrayrulewidth}
         & & \multicolumn{4}{c}{\textbf{per Distractor (Win$\uparrow$/Lose$\downarrow$)}} \\
        \cline{3-6}
        & & \multicolumn{2}{c|}{\textbf{Setting A}} & \multicolumn{2}{c}{\textbf{Setting B}} \\
        \cline{3-6}
        & & \shortstack{\\[0.1cm] Ours \\ (SFT)} & \shortstack{\\[0.1cm] Ours \\ (DPO)} 
        & \shortstack{\\[0.1cm] Ours \\ (SFT)}  & \shortstack{\\[0.1cm] Ours \\ (DPO)}  \\
        \hline
        \multirow{5}{*}{Python} & GPT-3.5 & 127/\textbf{129} & \textbf{145}/101 & 190/\textbf{198} & \textbf{198}/190 \\
            & GPT-4o & 158/\textbf{199} & \textbf{184}/156 & 178/\textbf{212} & \textbf{200}/185 \\
            &\citet{feng-2024-exploring} & 153/\textbf{157} & \textbf{164}/131 & 176/\textbf{223} & 196/\textbf{199} \\
            &\citet{hangetal-2024-mcqgen} & 110/\textbf{167} & \textbf{137}/120 & 179/\textbf{218} & \textbf{202}/198 \\
            &Human-Authored & 191/\textbf{199} & \textbf{207}/159 & \textbf{220}/175 & \textbf{217}/178 \\
        \hline
        \multirow{5}{*}{DB} & GPT-3.5 & 29/\textbf{32} & \textbf{28}/26 & 37/\textbf{48} & 34/\textbf{42} \\
            & GPT-4o & 40/\textbf{55} & \textbf{50}/41 & 38/\textbf{47} & \textbf{47}/39 \\
            &\citet{feng-2024-exploring} & 35/\textbf{47} & \textbf{41}/36 & \textbf{45}/41 & \textbf{45}/33 \\
            &\citet{hangetal-2024-mcqgen} & 34/\textbf{43} & \textbf{43}/25 & 37/\textbf{49} & 39/\textbf{52} \\            
            &Human-Authored & 25/\textbf{71} & 35/\textbf{54} &24/\textbf{53} & 31/\textbf{51} \\
        \hline
        \multirow{5}{*}{MLDL} & GPT-3.5 & 72/\textbf{73} & \textbf{68}/65 & \textbf{128}/115 & \textbf{150}/89 \\
            & GPT-4o & 104/\textbf{110} & \textbf{104}/91 & \textbf{135}/134 & \textbf{167}/99 \\
            &\citet{feng-2024-exploring} & 81/\textbf{90} & \textbf{84}/68 & \textbf{129}/123 & \textbf{150}/110 \\
            &\citet{hangetal-2024-mcqgen} & 57/\textbf{106} & 62/\textbf{83} & 109/\textbf{145} & \textbf{139}/113 \\
            &Human-Authored & 86/\textbf{130} & 81/\textbf{107} & 111/\textbf{141} & \textbf{127}/119 \\
        \Xhline{4\arrayrulewidth}
    \end{tabular}
    }
    \caption{Plausibility evaluation on distractor generators. Win/lose counts of our models (columns) against baselines (rows), averaged over two evaluations.}
    \label{tab:evaluation_DG_plausibility}
\end{table}

\paragraph{(1) Plausibility}
Table~\ref{tab:evaluation_DG_plausibility} summarizes the win/lose counts of our distractor generators against GPT models, \citet{feng-2024-exploring}, \citet{hangetal-2024-mcqgen} and human-authored distractors, as evaluated by our pairwise ranker (DPO-based). Our DPO model generated more plausible distractors than baseline models in most cases. Compared to human-authored distractors, our DPO model excelled in Python but underperformed in DB and MLDL. This discrepancy may be due to the underrepresentation of these subjects in our dataset, leading to limited exposure during training.

We assessed the benefit of augmenting the base MCQ dataset with synthetic distractors and automated ranking (i.e., the student choice dataset). Using only the base MCQ dataset for SFT and DPO led to a significant performance drop compared to using the whole student choice dataset, and no significant difference was observed between SFT and DPO (Appendix~\ref{sec:appendix_DG_ablation_study}). This highlights the importance of incorporating diverse chosen-rejected samples and sufficient distractors during training. 
Overall, the results demonstrate that our approach of creating the student choice dataset and employing DPO using this data effectively enhances distractor plausibility. 

We further examined the models' performance based on question type (i.e., code vs. statement). 
Our model outperformed GPT-3.5-turbo in generating plausible distractors for code type questions but was slightly less effective for statement type questions in Python and DB. In contrast, compared to GPT-4o, our model tended to perform better in statement type questions.
Detailed results are in Appendix~\ref{sec:appendix_DG_casestudy}.

\begin{table}[t]
    \centering
    \Large
    \renewcommand{\arraystretch}{1.35}
    \setlength{\arrayrulewidth}{0.15pt}
    \resizebox{\columnwidth}{!}{
    \begin{tabular}{l|ccccc|c}
        \Xhline{8\arrayrulewidth}
        & \multicolumn{5}{c|}{\textbf{\# of Selected Distractors $\uparrow$}} & \multicolumn{1}{c}{\textbf{DI $\uparrow$}}\\
        \cline{2-7}
        & \shortstack{Python \\[0.2pt]} & \shortstack{DB \\[3pt] } & \shortstack{MLDL \\[3pt]} & \shortstack{\\[2pt] Top \\ 50\%} & \shortstack{Low \\ 50\%} & \shortstack{Avg.\\[1pt]} \\
        \hline
        GPT-3.5 & 42 & \textbf{18} & 22 & 38 & 44 & 0.162  \\
        GPT-4o  & 14 & 5 & 26 & 22 & 23 & 0.119  \\
        \hline
        Ours (SFT) & 40 & 10 & 24 & 32 & 42 & 0.194 \\
        Ours (DPO) & \textbf{45} & 14 & \textbf{27} & \textbf{39} & \textbf{47} &\textbf{0.212} \\
        \Xhline{8\arrayrulewidth}
    \end{tabular}
    }
    \setlength{\arrayrulewidth}{0.4pt}
    \caption{Human evaluation on distractor generators.}
    \label{tab:evaluation_DG_human_evaluation}
\end{table}

\paragraph{(2) Human Evaluation}
Table~\ref{tab:evaluation_DG_human_evaluation} compares the frequency of distractors selected by students, showing that our DPO model generated more plausible distractors than GPT-4o across all subjects and outperformed GPT-3.5-turbo in all but one subject.
To evaluate whether the distractors have differing impacts based on students' proficiency levels, we divided the students into two groups---Top 50\% and Low 50\%---based on their average scores. The distractors generated by the DPO model were most frequently chosen by both groups.
These findings suggest that our model may effectively identify areas of confusion across varying proficiency levels as a versatile tool for a wide range of students.

Our DPO model achieved the highest discrimination index (DI) of 0.212, falling within the acceptable range of discrimination (0.21--0.24) \citep{kumar-2021-item}.
This indicates that the distractors generated by our model are better at differentiating between high-performing students and low-performing ones than the baseline models.
This is desirable because MCQs with a high DI can identify misconceptions and gaps in students' knowledge, and challenging MCQs can promote deeper learning.

Expert evaluation of our DPO model on \textit{clarity} and \textit{answerability} using a 5-point Likert scale showed that all metrics scored above 4, confirming that most distractors were clear enough to answer the question. Detailed results are provided in Appendix~\ref{sec:appendix_DG_human_evaluation}.

\paragraph{(3) Additional Evaluations}
We additionally evaluated the similarity between model-generated distractors and human-authored ones, as well as their validity. Our DPO model showed greater text similarity to human-authored distractors than GPT-3.5-turbo and GPT-4o. It also demonstrated higher validity compared to GPT-3.5-turbo, particularly excelling in questions that ask for incorrect statements.

Furthermore, we examined the similarity between model-generated distractors and the correct answer to assess the potential issue of distractors being too similar to the correct answer. Our analysis found no evidence that our models pose a particularly high risk to students because of this issue.
Detailed analyses can be found in Appendix~\ref{sec:appendix_DG_additional_evaluation}.

\paragraph{(4) Error Analysis}
We analyzed the suboptimal distractors generated by our model and identified several types of issues.
For code type questions, the distractors lacked variation in format, while for statement type questions, they were overly similar to the correct answers and failed to incorporate broader conceptual differences.
Examples of each type and future improvement strategies are detailed in Appendix~\ref{sec:appendix_DG_error_analysis}.

\begin{table}[t]
    \centering
    \small
    \renewcommand{\arraystretch}{1.4}
    \resizebox{\columnwidth}{!}{
    \begin{tabular}{ll|cc|cc} 
        \Xhline{3\arrayrulewidth}
         & & \multicolumn{4}{c}{\textbf{per Distractor (Win$\uparrow$/Lose$\downarrow$)}} \\
        \cline{3-6}
        & & \multicolumn{2}{c|}{\textbf{Setting A}} & \multicolumn{2}{c}{\textbf{Setting B}} \\
        \cline{3-6}
        & & \shortstack{\\Ours \\ (SFT)} & \shortstack{\\Ours \\ (DPO)} 
        & \shortstack{\\Ours \\ (SFT)}  & \shortstack{\\Ours \\ (DPO)}  \\
        \hline
        \multirow{2}{*}{Python} & GPT-3.5 & \textbf{71}/65 & \textbf{81}/56 & \textbf{100}/74 & \textbf{105}/59 \\
            & GPT-4o & 57/\textbf{78} & \textbf{74}/60 & \textbf{87}/53 & \textbf{95}/43 \\
        \hline
        \multirow{2}{*}{DB} & GPT-3.5 & \textbf{89}/67 & \textbf{88}/51 & \textbf{95}/64 & \textbf{107}/54 \\
            & GPT-4o & \textbf{93}/67 & \textbf{90}/53 & \textbf{103}/47 & \textbf{120}/33 \\
        \hline
        \multirow{2}{*}{MLDL} & GPT-3.5 & 59/\textbf{93} & 59/\textbf{82} & 71/\textbf{108} & 83/\textbf{96} \\
            & GPT-4o & 57/\textbf{100} & 67/\textbf{80} & 77/\textbf{98} & \textbf{90}/\textbf{90} \\
        \hline
        \multirow{2}{*}{English} & GPT-3.5 & \textbf{44}/\textbf{44} & 41/\textbf{42} & \textbf{78}/65 & \textbf{83}/61 \\
            & GPT-4o & 44/\textbf{46} & 39/\textbf{46} & \textbf{80}/64 & \textbf{84}/59 \\
        \Xhline{3\arrayrulewidth}
    \end{tabular}
    }
    \caption{Plausibility evaluation of distractor generators on four publicly available datasets (GPT-generated CS questions and a Korean high school English exam). For the English questions, plausibility was evaluated using GPT-4o due to its higher performance. Win/lose counts of our models (columns) against baselines (rows), averaged over two evaluations.}
    \label{tab:evaluation_DG_synthetic}
\end{table}

\begin{table}[t]
    \centering
    {\fontsize{9}{10}\selectfont 
    \renewcommand{\arraystretch}{1.4} 
    \setlength{\arrayrulewidth}{0.15pt}
    \begin{tabular}{l|ccc}
        \Xhline{6\arrayrulewidth}
        & \multicolumn{3}{c}{\textbf{Rank Accuracy $\uparrow$}} \\ \cline{2-4}
        & GPT-3.5 & GPT-4o & Ours (SFT)\\
        \hline
        English & 0.483 & \textbf{0.608} & 0.573 \\
        \Xhline{6\arrayrulewidth}
    \end{tabular}
    }
    \caption{Evaluation results on pairwise rankers for English  questions. The results were averaged over five generations for each model.}
    \label{tab:evaluation_PR_english}
\end{table}

\paragraph{(5) Generalizability}
To verify the generalizability of our approach beyond the base MCQ dataset and the CS domain, we conducted additional experiments on two publicly available datasets: (1) newly generated CS questions created using GPT-4o and (2) high school English exam questions.

For CS questions, we generated 100 MCQs per subject using GPT-4o and built a new student choice dataset to train a distractor generator.
The results in Table~\ref{tab:evaluation_DG_synthetic}, evaluated using our pairwise ranker, are consistent with those from the base MCQ dataset, reaffirming that plausibility improves with DPO over SFT.

For English questions, we used 88 questions from a South Korean high school exam\footnote{These MCQs are from the latest CSAT (Korean SAT), and the distractor selection rates were obtained from an online education platform specializing in CSAT preparation.} to train a pairwise ranker and a distractor generator. In Table~\ref{tab:evaluation_PR_english}, our pairwise ranker, despite limited training data, outperformed GPT-3.5-turbo and closely approached GPT-4o. Similarly, Table~\ref{tab:evaluation_DG_synthetic} shows that in Setting B, where more distractors were compared, our DPO model achieved higher plausibility than GPT models, reflecting the trends observed in CS subjects.


\section{Conclusion}
In this study, we proposed a pipeline for training a model to generate more plausible distractors for MCQs and demonstrated its effectiveness across computer science subjects. 
We trained the pairwise ranker to evaluate the relative plausibility of distractors, and used this to create the student choice dataset where distractors for each question are ranked by plausibility. 
From this dataset, we created chosen-rejected pairs of distractors to train the distractor generator using DPO.
Our models outperformed GPT and other baseline models and performed comparably to humans in various metrics, including pairwise rank accuracy and distractor plausibility.
We believe that our work can advance automated educational tools, contributing to more adaptive and effective learning environments.

\section*{Limitations}
The models presented in this study have the following limitations.
First, the pairwise ranker's method of comparing distractors pairwise significantly increases the number of combinations and requires substantial computing resources due to the need for generating reasoning. A listwise approach using an encoder-decoder structure could be explored as a solution \citep{yoon-etal-2024-listt5}.

Second, the distractor generator occasionally produces invalid distractors, necessitating review by human experts or high-performing LLMs (e.g., GPT-4o) to accurately evaluate students' knowledge. To address this limitation, future work could include an additional supervision phase, such as integrating feedback loops with other models or applying constraints like \textit{Counterfactual Contrastive Decoding} \citep{qu-etal-2024-unsupervised}.

Finally, our method focuses on generating difficult distractors, but there are instances where adjusting the difficulty level of MCQs to suit the needs of the target students is necessary. While our pairwise ranker can be utilized to select distractors with varying degrees of plausibility, future work could explore more direct approaches, such as incorporating student knowledge tracing or adaptive decoding, to address this challenge \citep{cui-sachan-2023-adaptive}.

\section*{Acknowledgments}
This work was supported by the National Research Foundation of Korea (NRF) grants funded by the Korean government (MSIT) (RS-2024-00333484, RS-2024-00414981). It was also supported by Elice, Inc., which also provided the proprietary datasets.

\bibliography{anthology, custom}

\appendix

\section{Pairwise Ranker}

\subsection{Prompt for Pairwise Ranker} \label{sec:appendix_PR_prompt}
The instruction prompts for the pariwise ranker are in Table~\ref{tab:appendix_PR_prompt_reasoning} (Reasoning),~\ref{tab:appendix_PR_prompt_rubric} (Rubric),~\ref{tab:appendix_PR_prompt_g-eval} (G-Eval),~\ref{tab:appendix_PR_prompt_discussion} (Discussion) and~\ref{tab:appendix_PR_prompt_scarlatos} \citep{scarlatos-etal-2024-improving}. We used the same prompt (Reasoning) with GPT models and ours (SFT, DPO).

\subsection{SFT and DPO Settings for Pairwise Ranker} \label{sec:appendix_PR_settings}
The pairwise ranker model was trained using Mistral-7B-Instruct-v0.2\footnote{This model is distributed under the Apache 2.0 license. \href{https://huggingface.co/mistralai/Mistral-7B-Instruct-v0.2}{https://huggingface.co/mistralai/Mistral-7B-Instruct-v0.2}} with 4-bit quantization and fine-tuned using LoRA. For SFT, the learning rate was set to 2e-4 and the model was trained for 5 epochs. For DPO, the learning rate was set to 1e-6, also trained for 5 epochs. These hyperparameters were selected as they allowed stable training without overfitting while preserving the quality of the DPO output. 
SFT took approximately 2 hours, and DPO took about 1 hour on an NVIDIA A6000 GPU.
\citet{scarlatos-etal-2024-improving} model was reproduced for baseline comparison using the same model and DPO settings as above.

\subsection{Prompt for Generating Pairwise Ranker Training Data} \label{sec:appendix_PR_training}
The instruction prompt for generating pairwise ranker training data is in Table~\ref{tab:appendix_PR_training}. To enhance the diversity of expressions and reasoning used in the samples, two reasoning examples are generated for each pair—one with temperature set to 0 and the other to 1—using GPT-4o.

\subsection{Consistency in Rank Prediction} \label{sec:appendix_PR_generation_attempts}
Table~\ref{tab:evaluation_PR_generation_attemps} demonstrates that our pairwise ranker exhibits relatively robust to positional bias.
In comparison to GPT-3.5-turbo, which required an average of more than two attempts to produce consistent results when the input order was altered, our DPO model was able to achieve consistent results with significantly fewer attempts.
Additionally, our DPO model slightly outperformed GPT-4o by requiring fewer average generation attempts.

\subsection{Plausibility Factors} \label{sec:appendix_PR_factors}
We used GPT-4o to summarize and categorize reasoning samples where our pairwise ranker accurately predicted the rankings on the test set, and selected six representative examples per question type.
Definitions for each category are in Table~\ref{tab:appendix_PR_factors_code} (Code Type) and~\ref{tab:appendix_PR_factors_statement} (Statement Type).

\subsection{Human Evaluation} \label{sec:appendix_PR_human_evaluation}
\paragraph{Recruitment}
We conducted a survey with three Master's degree students who voluntarily expressed their willingness to participate in this experiment. The survey was designed to begin only after they agreed to provide their results for research purposes and acknowledged the precautions via an online form. The experiment lasted approximately 90 minutes, and participants were compensated above the standard hourly wage for the time they participated.
The entire process of human evaluation was conducted following procedures approved by the IRB committee of our university.

\paragraph{Survey Form}
The reasoning quality of our pairwise ranker was evaluated on a 5-point Likert scale based on the following criteria:
\begin{itemize}
    \item \textbf{Logical Reasoning:} Whether the reasoning process is logical.
    \item \textbf{Student Understanding:} Whether the reasoning effectively understands students' misconceptions or problem-solving processes.
    \item \textbf{Knowledge Accuracy:} Whether the reasoning is based on accurate and error-free knowledge.
    \item \textbf{Choice Agreement:} Whether the evaluator agrees with the model's final choice.
\end{itemize}

An example of the survey form is presented in Table~\ref{tab:appendix_PR_human_evaluation}.

\begin{table}[t]
    \centering
    \Large
    \renewcommand{\arraystretch}{1.5}
    \setlength{\arrayrulewidth}{0.15pt}
    \resizebox{\columnwidth}{!}{
    \begin{tabular}{l|>{\centering\arraybackslash}p{1.7cm}>{\centering\arraybackslash}p{1.7cm}>{\centering\arraybackslash}p{1.7cm}>{\centering\arraybackslash}p{1.7cm}}
        \Xhline{8\arrayrulewidth}
        & \multicolumn{4}{c}{\textbf{Generation Attempts per Question $\downarrow$}} \\ \cline{2-5}
        & Python & DB & MLDL & Avg. \\
        \hline
        GPT-3.5 (Reasoning) & 2.491 & 2.482 & 2.427 & 2.467 \\
        GPT-4o (Reasoning)  & 1.753 & \textbf{1.699} & \textbf{1.699} & 1.717 \\
        GPT-4o (Rubric) & 2.306 & 2.316 & 2.212 & 2.278 \\
        GPT-4o (G-Eval) & 5.303 & 5.193 & 5.150 & 5.215 \\
        \hline
        Ours (SFT, Sep.) & 1.708 & 1.718 & 1.771 & 1.732 \\
        Ours (SFT, Comb.) & 1.685 & 1.715 & 1.740 & 1.713 \\
        Ours (DPO, Comb.) & \textbf{1.650} & 1.725 & 1.740 & \textbf{1.705} \\
        \hline
        Ours (SFT w/o Reasoning) & 2.013 & 2.034 & 2.036 & 2.028 \\
        \Xhline{8\arrayrulewidth}
    \end{tabular}
    }
    \setlength{\arrayrulewidth}{0.4pt}
    \caption{Consistency evaluation results on pairwise rankers. The results were averaged over five generations for each model.}
    \label{tab:evaluation_PR_generation_attemps}
\end{table}

\subsection{Ablation Study} \label{sec:appendix_PR_ablation_study}
The instruction prompt used for the ablation study (w/o Reasoning) is in Table~\ref{tab:appendix_PR_ablation_study}, and the training settings are identical to those of our pairwise ranker training setup (Appendix~\ref{sec:appendix_PR_settings}).

\subsection{Error Analysis} \label{sec:appendix_PR_error_analysis}
Our pairwise ranker exhibited the following three types of errors:

First, our model tended to incorrectly judge implausible mistakes as plausible—errors that real students would not typically make. 
For example, in the process of calculating the output of Python code, the model incorrectly deemed `unrealistic reasoning' or `mistakes in obvious calculations' as plausible, even though such errors would be unlikely for actual students to make based on common sense.

Second, our model struggled with reasoning when encountering unfamiliar questions that were insufficiently represented in the training data. 
This issue was particularly evident in subjects like DB and MLDL, where the training set was relatively small and shared few similar concepts or questions with the test set.

Lastly, in questions requiring the selection of an \textit{incorrect} option, there were cases where our model’s final ranking was correct, but its reasoning was flawed. 
Instead of identifying why each option seemed more incorrect to the students, the model mistakenly focused on determining which option was more correct.

To improve the pairwise ranker, future work should focus on enabling the model to learn \textit{common student misconceptions} for better reasoning and prediction and enhancing the inference process to clearly recognize \textit{question requirements}.


\section{Distractor Generator}

\subsection{Prompt for Distractor Generator} \label{sec:appendix_DG_prompt}
The instruction prompt for our distractor generator is in Table~\ref{tab:appendix_DG_prompt}. We used the same prompt for both GPT models and ours. However, we instructed the GPT models to generate outputs in \textit{JSON} format for stability reasons.

The instruction prompt for the kNN approach proposed by \citet{feng-2024-exploring} is presented in Table~\ref{tab:appendix_DG_knn_prompt}. Following the method outlined in the paper, the target question and answer were encoded using the SBERT encoder \citep{reimers-gurevych-2019-sentence}, MPNet\footnote{\href{https://huggingface.co/sentence-transformers/all-mpnet-base-v2}{https://huggingface.co/sentence-transformers/all-mpnet-base-v2}}, and the top-3 most similar items based on cosine similarity were extracted from the question pool (training set) and used as in-context examples.

\subsection{SFT and DPO Settings for Distractor Generator} \label{sec:appendix_DG_settings}
The distractor generator model was trained using Mistral-7B-Instruct-v0.2 with 4-bit quantization and fine-tuned using LoRA. For SFT, the learning rate was set to 2e-4 and the model was trained for 2 epochs. For DPO, the learning rate was set to 1e-5, trained for 3 epochs. 
These hyperparameters were determined as a result of finding a setup that avoids overfitting while ensuring no issues with the quality of the DPO output.
SFT and DPO took approximately 3 hours on an NVIDIA A6000 GPU.

As briefly mentioned in \S{\ref{sec:methods_subsection_distractor_generator}}, in addition to the chosen-rejected sample pairing method described in the main text, another setting employs a method similar to a \textit{sliding window} for pairing. 
In this setting, all distractor candidates are sorted in descending order and grouped into non-overlapping windows of size $n$. 
For example, if there are six candidates and $n$ is 2, a total of three windows are created. 
Pairwise combinations between these windows are then used to create chosen-rejected samples. 
A model trained with DPO using these samples showed no significant performance difference compared to the model described in the main text. 
The plausibility evaluation results for this model are provided in Table~\ref{tab:appendix_DG_slide_window_plausibility}.

\begin{table}[t]
    \centering
    \Large
    \renewcommand{\arraystretch}{1.4}
    \resizebox{\linewidth}{!}{
    \begin{tabular}{ll|c|c|c|c} 
        \Xhline{4\arrayrulewidth}
        & & \multicolumn{2}{c|}{\textbf{\shortstack{\\[0.1cm] per Question \\ (Win$\uparrow$/Tie/Lose$\downarrow$)}}} & \multicolumn{2}{c}{\textbf{\shortstack{\\[0.1cm] per Distractor \\ (Win$\uparrow$/Lose$\downarrow$)}}} \\
        \cline{3-6}
        & & \shortstack{\textbf{Setting A}} & \shortstack{\textbf{Setting B}} & \shortstack{\textbf{Setting A}} & \shortstack{\textbf{Setting B}} \\ 
        \cline{3-6}
        & & \shortstack{\\[0.1cm] Ours \\ (DPO, \\  window)} & \shortstack{\\[0.1cm] Ours \\ (DPO, \\  window)} & \shortstack{\\[0.1cm] Ours \\ (DPO, \\  window)} & \shortstack{\\[0.1cm] Ours \\ (DPO, \\  window)} \\
        \hline
        \multirow{2}{*}{Python} & GPT-3.5 & \multicolumn{1}{c|}{\textbf{21}/13/18} & \multicolumn{1}{c|}{\textbf{30}/1/20} & \multicolumn{1}{c|}{\textbf{140.5}/110.5} & \multicolumn{1}{c}{\textbf{216}/173} \\
        & GPT-4o & \multicolumn{1}{c|}{21/9/\textbf{22}} & \multicolumn{1}{c|}{\textbf{22}/11/18} & \multicolumn{1}{c|}{\textbf{186}/162} & \multicolumn{1}{c}{\textbf{200}/187} \\
        \hline
        \multirow{2}{*}{DB} & GPT-3.5 & \multicolumn{1}{c|}{4/\textbf{6}/3} & \multicolumn{1}{c|}{\textbf{6}/1/4} & \multicolumn{1}{c|}{\textbf{33.5}/25.5} & \multicolumn{1}{c}{\textbf{42.5}/33.5} \\
        & GPT-4o & \multicolumn{1}{c|}{\textbf{6}/2/5} & \multicolumn{1}{c|}{\textbf{5}/4/2} & \multicolumn{1}{c|}{\textbf{48.5}/44.5} & \multicolumn{1}{c}{\textbf{50.5}/40.5} \\
        \hline
        \multirow{2}{*}{MLDL} & GPT-3.5 & \multicolumn{1}{c|}{7/\textbf{12}/\textbf{12}} & \multicolumn{1}{c|}{13/2/\textbf{15}} & \multicolumn{1}{c|}{60.5/\textbf{72.5}} & \multicolumn{1}{c}{\textbf{136.5}/107.5} \\
        & GPT-4o & \multicolumn{1}{c|}{12/6/\textbf{14}} & \multicolumn{1}{c|}{21/1/10} & \multicolumn{1}{c|}{95.5/\textbf{102.5}} & \multicolumn{1}{c}{\textbf{165}/107} \\
        \Xhline{4\arrayrulewidth}
    \end{tabular}
    }
    \caption{Plausibility evaluation on the distractor generator, DPO with sliding window setting (Appendix~\ref{sec:appendix_DG_settings}).}
    \label{tab:appendix_DG_slide_window_plausibility}
\end{table}

\begin{table}[t]
    \centering
    \Large
    \renewcommand{\arraystretch}{1.3}
    \resizebox{\columnwidth}{!}{
    \begin{tabular}{ll|cc|cc} 
        \Xhline{4\arrayrulewidth}
         & & \multicolumn{4}{c}{\textbf{per Question (Win$\uparrow$/Tie/Lose$\downarrow$)}} \\
        \cline{3-6}
        & & \multicolumn{2}{c|}{\textbf{Setting A}} & \multicolumn{2}{c}{\textbf{Setting B}} \\
        \cline{3-6}
        & & \shortstack{\\[0.1cm] Ours \\ (SFT)} & \shortstack{\\[0.1cm] Ours \\ (DPO)} 
        & \shortstack{\\[0.1cm] Ours \\ (SFT)}  & \shortstack{\\[0.1cm] Ours \\ (DPO)}  \\
        \hline
        \multirow{4}{*}{Python} & GPT-3.5 & 17/13/\textbf{22} & \textbf{23}/15/14 & 21/7/\textbf{22} & \textbf{26}/1/23 \\
            & GPT-4o & 12/15/\textbf{25} & \textbf{24}/12/16 & 19/6/\textbf{26} & \textbf{24}/4/22 \\
            &\citet{feng-2024-exploring} & 14/\textbf{21}/17 & \textbf{21}/15/15 & 16/5/\textbf{28} & \textbf{25}/2/21 \\
            &Human-Authored & \textbf{21}/11/20 & \textbf{27}/10/15 & \textbf{31}/5/15 & \textbf{26}/4/20 \\            
        \hline
        \multirow{4}{*}{DB} & GPT-3.5 & 3/3/\textbf{7} & \textbf{6}/2/5 & 4/1/\textbf{5} & \textbf{6}/1/4 \\
            & GPT-4o & 4/1/\textbf{8} & 4/3/\textbf{6} & 5/0/\textbf{6} & \textbf{6}/0/5 \\
            &\citet{feng-2024-exploring} & 3/3/\textbf{7} & \textbf{6}/3/4 & \textbf{6}/1/3 & \textbf{6}/0/4 \\
            &Human-Authored & 1/1/\textbf{11} & 4/0/\textbf{9} & 5/1/\textbf{7} & 4/2/\textbf{7} \\               
        \hline
        \multirow{4}{*}{MLDL} & GPT-3.5 & 9/\textbf{12}/10 & 12/\textbf{13}/6 & 12/1/\textbf{17} & \textbf{18}/2/10 \\
            & GPT-4o & 11/4/\textbf{17} & \textbf{12}/7/\textbf{12} & 13/4/\textbf{15} & \textbf{19}/7/5 \\
            &\citet{feng-2024-exploring} & 11/6/\textbf{14} & \textbf{15}/6/9 & \textbf{17}/0/15 & \textbf{18}/3/10 \\
            &Human-Authored & 11/4/\textbf{17} & 9/7/\textbf{16} & 11/3/\textbf{18} & \textbf{15}/2/4 \\   
        \Xhline{4\arrayrulewidth}
    \end{tabular}
    }
    \caption{Plausibility evaluation on distractor generators.}
    \label{tab:evaluation_DG_plausibility_per_question}
\end{table}

\subsection{Plausibility Evaluation}
\label{sec:appendix_DG_casestudy}
\paragraph{per Question}
The results of the plausibility evaluation analyzed from a per-question perspective are presented in Table~\ref{tab:evaluation_DG_plausibility_per_question} (compare with Table~\ref{tab:evaluation_DG_plausibility}).

\paragraph{Case Study}
The analysis of plausibility results based on question types (Code/Statement) is provided in Table~\ref{tab:appendix_evaluation_DG_case_study}. 
A summary of the case study results is as follows:

First, our model generates more plausible distractors for code type questions compared to GPT-3.5-turbo. The distractors generated by the latter were either significantly different from the correct answer or included code syntax that does not actually exist. 
On the other hand, for the statement type questions, GPT-3.5-turbo demonstrated higher plausibility only in the cases of Python and DB.
This was because its distractors included more diverse knowledge or additional conditions, while our model seemed to construct distractors with relatively limited scope of knowledge, possibly due to the small training dataset.

Next, our model exhibited higher plausibility in the statement type compared to GPT-4o. When compared with the validity results in Appendix~\ref{sec:appendix_DG_additional_evaluation} (`Statement'), it can be seen that GPT-4o generated more obvious statements, resulting in a lower risk of invalid distractors but making the difficulty level lower. 
For the code type, both models generated distractors that were not far from the correct answer. 
However, in the case of Python, the distractors generated by our model were slightly less plausible than those of GPT-4o, likely because the latter made better use of partial errors in the code.

\begin{table}[t]
    \centering
    \normalsize
    \renewcommand{\arraystretch}{1.4}
    \resizebox{\linewidth}{!}{
    \begin{tabular}{ll|cc|cc} 
        \Xhline{4\arrayrulewidth}
        & & \multicolumn{2}{c|}{\textbf{\shortstack{\\[0.1cm] per Question \\ (Win$\uparrow$/Lose$\downarrow$)}}} & \multicolumn{2}{c}{\textbf{\shortstack{\\[0.1cm] per Distractor \\ (Win$\uparrow$/Lose$\downarrow$)}}} \\
        \cline{3-6}
        & & Code & State. & Code & State. \\
        \hline
        \multirow{2}{*}{Python} & GPT-3.5 & \multicolumn{1}{c}{\textbf{12}/5} & \multicolumn{1}{c|}{14/\textbf{18}} & \multicolumn{1}{c}{\textbf{86.5}/52.5} & \multicolumn{1}{c}{111.5/\textbf{137.5}} \\
        & GPT-4o & \multicolumn{1}{c}{6/\textbf{11}} & \multicolumn{1}{c|}{\textbf{18}/11} & \multicolumn{1}{c}{58.5/\textbf{70.5}} & \multicolumn{1}{c}{\textbf{142}/115} \\
        \hline
        \multirow{2}{*}{DB} & GPT-3.5 & \multicolumn{1}{c}{\textbf{4}/0} & \multicolumn{1}{c|}{2/\textbf{4}} & \multicolumn{1}{c}{\textbf{22.5}/8.5} & \multicolumn{1}{c}{11.5/\textbf{33.5}} \\
        & GPT-4o & \multicolumn{1}{c}{\textbf{2}/\textbf{2}} & \multicolumn{1}{c|}{\textbf{4}/3} & \multicolumn{1}{c}{\textbf{20}/16} & \multicolumn{1}{c}{\textbf{27}/23} \\
        \hline
        \multirow{2}{*}{MLDL} & GPT-3.5 & \multicolumn{1}{c}{-} & \multicolumn{1}{c|}{\textbf{18}/10} & \multicolumn{1}{c}{-} & \multicolumn{1}{c}{\textbf{150}/89} \\
        & GPT-4o & \multicolumn{1}{c}{-} & \multicolumn{1}{c|}{\textbf{19}/5} & \multicolumn{1}{c}{-} & \multicolumn{1}{c}{\textbf{167}/99} \\
        \Xhline{4\arrayrulewidth}
    \end{tabular}
    }
    \caption{Plausibility evaluation on the distractor generator, categorized by question type. This table further details the results from Table~\ref{tab:evaluation_DG_plausibility} and~\ref{tab:evaluation_DG_plausibility_per_question}, Setting B, Ours (DPO).}
    \label{tab:appendix_evaluation_DG_case_study}
\end{table}

\subsection{Human Evaluation} \label{sec:appendix_DG_human_evaluation}
\paragraph{Recruitment}
We conducted the evaluation with 15 college students who voluntarily agreed to participate. The test was conducted online, and participants were allowed to begin the test only after agreeing to the instruction stating that their results would be provided for research purposes. The experiment took approximately 60 minutes, and participants were compensated with a reward above the standard hourly wage for their time. The entire human evaluation process was conducted in accordance with the procedures approved by the IRB committee of our university.

\paragraph{Test Form}
Each question allows for multiple selections (e.g., Select all the correct/incorrect ...) and includes one distractor generated by each model, along with `None of the above' as the final option. 
To mitigate unintended effects on the selection rate of distractors when the actual correct answer is included, two versions of each question were created: one with the correct answer included and one without. 
These versions were randomly distributed. For analysis, the results from both versions were integrated.

\begin{table}[t]
    \centering
    {\fontsize{9}{10}\selectfont 
    \renewcommand{\arraystretch}{1.4} 
    \setlength{\arrayrulewidth}{0.15pt} 
    \begin{tabular}{l|ccc}
        \Xhline{8\arrayrulewidth}
        & \multicolumn{3}{c}{\textbf{5-point Likert Scale $\uparrow$}} \\ 
        \cline{2-4}
        & Python & DB & MLDL \\
        \hline
        Clairity & 4.418 & 4.282 & 4.373 \\
        Answerability & 4.414 & 4.264 & 4.382 \\
        \Xhline{8\arrayrulewidth}
    \end{tabular}
    } 
    \caption{Evaluation results on the clarity and answerability of distractors generated by our DPO model.}
    \label{tab:appendix_DG_answerability}
\end{table}

\begin{table}[t]
    \centering
    \Large
    \renewcommand{\arraystretch}{1.4}
    \resizebox{\columnwidth}{!}{
    \begin{tabular}{l|ccc|ccc} 
        \Xhline{4\arrayrulewidth}
        & \multicolumn{3}{c|}{\textbf{sBLEU $\uparrow$}} 
        & \multicolumn{3}{c}{\textbf{BERTScore $\uparrow$}} \\ \cline{2-7}
        & Python & DB & MLDL 
        & Python & DB & MLDL  \\ \hline
        GPT-3.5       & 12.572 & 16.794 & 10.133 
                      & 0.879  & 0.893  & 0.873   \\
        GPT-4o        & 15.387 & 24.752 & 16.120 
                      & 0.893  & \textbf{0.912}  & \textbf{0.882}   \\
        Mistral       & 11.192 & 14.642 & 10.850
                      & 0.859  & 0.872  & 0.863 \\
        \hline
        Ours (SFT)    & 16.859 & 20.892 & 14.752 
                      & 0.894  & 0.897  & 0.876  \\
        Ours (DPO)    & \textbf{18.313} & \textbf{26.322} & \textbf{16.476} 
                      & \textbf{0.896}  & 0.906  & 0.881  \\ 
        \Xhline{4\arrayrulewidth}
    \end{tabular}
    }
    \caption{Similarity between model-generated and human-authored distractors.}
    \label{tab:appendix_DG_evaluation_token_match}
\end{table}

\begin{table*}[t]
    \centering
    \small
    \renewcommand{\arraystretch}{1.4}
    \resizebox{\textwidth}{!}{
    \begin{tabular}{l|ccccccc} 
        \Xhline{3\arrayrulewidth}
         & \multicolumn{7}{c}{\textbf{Similarity}}  \\
        \cline{2-8}
        & Human	& GPT-3.5 & GPT-4o & \citet{feng-2024-exploring} & \citet{hangetal-2024-mcqgen} & Ours (SFT) & Ours (DPO) \\ 
        \hline
        Python & 0.53 (0.04) & 0.54 (0.05) & 0.60 (0.05) & 0.58 (0.04) & 0.59 (0.05) & 0.57 (0.06) & 0.59 (0.06) \\
        DB & 0.55 (0.05) & 0.55 (0.05) & 0.60 (0.05) & 0.58 (0.04) & 0.59 (0.05) & 0.59 (0.06) & 0.61 (0.06) \\
        MLDL & 0.53 (0.04) & 0.47 (0.05) & 0.51 (0.05) & 0.52 (0.04) & 0.52 (0.05) & 0.50 (0.05) & 0.53 (0.05) \\
        \Xhline{3\arrayrulewidth}
    \end{tabular}
    }
    \caption{Similarity between model-generated distractors and correct answers. Numbers in parentheses represent variance.}
    \label{tab:appendix_DG_additional_similarity}
\end{table*}

\paragraph{DI}
To analyze the DI of a specific model, it is necessary to assume that each item consists solely of options generated by that model. Therefore, we restructured the test results by treating each distractor generated by a model as a separate test item that determines `whether the corresponding statement (distractor) is true or false'. In other words, we assumed that all students took multiple independent tests, each consisting of items created exclusively with distractors from a single model.
When grading, if a student chose the distractor generated by the model, the item was considered incorrect; otherwise, it was considered correct. The cutoff for dividing students into high and low groups was set at the top and bottom 27\%, and the DI calculation formula was also in line with previous studies \citep{mahjabeen-2017-difficulty, rezigalla-2024-item}.

\paragraph{Clairity and Answerability}
We concern that excessively ambiguous distractors could hinder educational assessment. To verify whether such issues exist, we asked 11 experts to evaluate the clarity and answerability  of distractors generated by our DPO model on a 5-point Likert scale. The results, averaged across a total of 40 questions (Python: 20, DB: 10, MLDL: 10) for each subject, are in Table~\ref{tab:appendix_DG_answerability}.
These results confirm that the potential side effects of high plausibility, which were a concern, do not appear to be present in our model.

\subsection{Additional Evaluations}
\label{sec:appendix_DG_additional_evaluation}
\paragraph{Similarity Between Model-Generated and Human-Authored Distractors}
Table~\ref{tab:appendix_DG_evaluation_token_match} presents the similarity evaluation results for the distractor generator.
sBLEU\footnote{\href{https://github.com/mjpost/sacrebleu}{https://github.com/mjpost/sacrebleu}} and BERTScore\footnote{\href{https://github.com/Tiiiger/bert_score}{https://github.com/Tiiiger/bert\_score}} were used as the text similarity metrics.
For sBLEU, the `smooth\_method' was set to `exp', and the default parameters were used for BERTScore.
In terms of sBLEU, our model (DPO) generates distractors that are most similar to human-authored ones across the majority of subjects.

\paragraph{Similarity Between Model-Generated Distractors and Correct Answers}
We evaluated the semantic similarity between distractors and correct answers using OpenAI’s \textit{text-embedding-3-small} model. Cosine similarity scores were computed between correct answers and distractors generated by each model.

As shown in Table~\ref{tab:appendix_DG_additional_similarity}, similarity scores are largely consistent across models, with no evidence that our SFT or DPO models generate distractors that are excessively similar to correct answers. For instance, the average similarity score of human-authored distractors (0.53) is comparable to that of our DPO model (0.59) for Python.

\begin{itemize}
    \item \textbf{Question:} "Choose the incorrect statement."
    \item \textbf{Correct Answer:} "If an if statement's condition is False, its content executes before the else statement."
    \item \textbf{Distractor 1 (similarity = 0.59):} "An if-else statement can include multiple conditions using elif in some languages."
    \item \textbf{Distractor 2 (similarity = 0.53):} "An if statement can exist without an else statement in most languages."
\end{itemize}

A 0.06 difference in similarity score does not indicate a meaningful impact on quality, as the first distractor is not noticeably more similar to the correct answer than the second.

\begin{table*}[t]
    \centering
    \normalsize
    \renewcommand{\arraystretch}{1.4}
    \resizebox{\textwidth}{!}{
    \begin{tabular}{l|ccc|ccc|ccc|ccc} 
        \Xhline{4\arrayrulewidth}
         & \multicolumn{12}{c}{\textbf{Validity $\uparrow$}} \\
        \cline{2-13}
         & \multicolumn{3}{c|}{\textbf{Correct}} & \multicolumn{3}{c|}{\textbf{Incorrect}} & \multicolumn{3}{c|}{\textbf{Code}} & \multicolumn{3}{c}{\textbf{Statement}} \\
        \cline{2-13}
         & Python & DB & MLDL 
         & Python & DB & MLDL 
         & Python & DB & MLDL 
         & Python & DB & MLDL \\
        \hline
        GPT-3.5 & 0.883 & 0.571 & 0.938 & 0.400 & \textbf{1.000} & 0.426 & 0.877 & 0.630 & - & 0.588 & 0.630 & 0.684 \\
        GPT-4o & \textbf{0.938} & \textbf{1.000} & 0.902 & \textbf{0.967} & 0.917 & \textbf{0.956} & \textbf{0.912} & 0.917 & - & \textbf{0.970} & \textbf{0.963} & \textbf{0.927} \\
        Mistral (w/o $T$) & 0.839 & 0.227 & 0.820 & 0.233 & 0.917 & 0.370 & 0.815 & 0.917 & - & 0.485 & 0.357 & 0.604 \\
        Mistral (w/ $T$) & 0.903 & 0.364 & 0.822 & 0.265 & 0.750 & 0.378 & 0.891 & 0.750 & - & 0.500 & 0.464 & 0.600 \\
        \hline
        Ours (SFT) & 0.874 & 0.905 & 0.765 & 0.902 & \textbf{1.000} & 0.844 & 0.842 & \textbf{1.000} & - & 0.909 & 0.926 & 0.802 \\
        Ours (DPO) & 0.839 & 0.905 & 0.627 & 0.850 & 0.917 & 0.800 & 0.875 & 0.917 & - & 0.825 & 0.889 & 0.708 \\
        \hline
        Ablation (SFT) & 0.783 & 0.778 & 0.608 & 0.733 & 0.810 & 0.822 & 0.717 & 0.833 & - & 0.788 & 0.778 & 0.708 \\
        Ablation (DPO) & 0.848 & 0.722 & 0.627 & 0.717 & 0.905 & 0.733 & 0.811 & 0.750 & - & 0.788 & 0.852 & 0.677 \\
        \Xhline{4\arrayrulewidth}
    \end{tabular}
    }
    \caption{Validity evaluation on distractor generators. Mistral is a model that has not been fine-tuned, w/o $T$ is the result of using a prompt that generates distractors directly without specifying the distractor type, and w/ $T$ is the result using the same prompt as Ours.}
    \label{tab:appendix_DG_evaluation_validity}
\end{table*}

\paragraph{Validity}
Validity refers to whether the distractors are indeed incorrect options for the question.
We assessed validity by categorizing questions as either Correct/Incorrect or Code/Statement.
`Correct' and `Incorrect' refer to question types where the task is to select the correct or incorrect statement, respectively.
`Code' type questions involve cases where the answer (and distractors) take the form of filling in blanks or matching outputs in code.
`Statement' refers to questions composed of explanatory statements about a concept.

Table~\ref{tab:appendix_DG_evaluation_validity} shows the proportion of valid distractors generated by each model according to the type of question. Our models demonstrate stable validity across various question types, significantly outperforming GPT-3.5-turbo and pre-trained Mistral. This highlights the importance of the proposed methodology—first generating the type ($T$) such as `Correct/Incorrect knowledge'—in enhancing validity. 

Studies have shown that LLMs perform poorly on tasks involving negation \citep{varshney-2024-investigating}, and in a similar vein, GPT-3.5-turbo and Mistral show significantly lower validity when generating distractors for question types that require selecting an incorrect option (in `Incorrect' type, the distractors should actually represent correct knowledge, but these models mostly generated distractors with incorrect knowledge). 
However, after going through SFT and DPO, the proportion of valid distractors generated for such types greatly increases, indicating that the proposed methodology in this study (first generating types such as `Correct/Incorrect knowledge') plays an important role in improving validity. 
Meanwhile, there is a slight decrease in validity after DPO compared to SFT, which appears to be a trade-off arising from the process of creating more confusing distractors.

\begin{table}[t]
    \centering
    \huge
    \renewcommand{\arraystretch}{1.4}
    \resizebox{\linewidth}{!}{
    \begin{tabular}{ll|cc|cc} 
        \Xhline{8\arrayrulewidth}
        & & \multicolumn{2}{c|}{\textbf{\shortstack{\\[0.1cm] per Question \\ (Win$\uparrow$/Tie/Lose$\downarrow$)}}} & \multicolumn{2}{c}{\textbf{\shortstack{\\[0.1cm] per Distractor \\ (Win$\uparrow$/Lose$\downarrow$)}}} \\
        \cline{3-6}
        & & \shortstack{\\[0.1cm] Ablation \\ (SFT)} & \shortstack{\\[0.1cm] Ablation \\ (DPO)} & \shortstack{\\[0.1cm] Ablation \\ (SFT)} & \shortstack{\\[0.1cm] Ablation \\ (DPO)} \\
        \hline
        \multirow{2}{*}{Python} & GPT-3.5 & \multicolumn{1}{c}{15/14/\textbf{23}} & \multicolumn{1}{c|}{\textbf{20}/13/19} & \multicolumn{1}{c}{98.5/\textbf{121.5}} & \multicolumn{1}{c}{115.5/\textbf{122}} \\
        & GPT-4o & \multicolumn{1}{c}{14/11/\textbf{27}} & \multicolumn{1}{c|}{18/9/\textbf{25}} & \multicolumn{1}{c}{122/\textbf{179}} & \multicolumn{1}{c}{135.5/\textbf{184.5}} \\
        \hline
        \multirow{2}{*}{DB} & GPT-3.5 & \multicolumn{1}{c}{\textbf{5}/\textbf{5}/3} & \multicolumn{1}{c|}{3/4/\textbf{6}} & \multicolumn{1}{c}{\textbf{29}/21} & \multicolumn{1}{c}{25.5/\textbf{26.5}} \\
        & GPT-4o & \multicolumn{1}{c}{\textbf{7}/1/5} & \multicolumn{1}{c|}{4/4/\textbf{5}} & \multicolumn{1}{c}{\textbf{46.5}/36} & \multicolumn{1}{c}{42/\textbf{44}} \\
        \hline
        \multirow{2}{*}{MLDL} & GPT-3.5 & \multicolumn{1}{c}{8/10/\textbf{13}} & \multicolumn{1}{c|}{9/9/\textbf{13}} & \multicolumn{1}{c}{55/\textbf{70.5}} & \multicolumn{1}{c}{52/\textbf{66}} \\
        & GPT-4o & \multicolumn{1}{c}{\textbf{13}/7/12} & \multicolumn{1}{c|}{\textbf{15}/8/9} & \multicolumn{1}{c}{87.5/\textbf{104.5}} & \multicolumn{1}{c}{\textbf{95.5}/82} \\
        \Xhline{8\arrayrulewidth}
    \end{tabular}
    }
    \caption{Ablation study on our distractor generator. The evaluation setup is the same as Setting A in Table~\ref{tab:evaluation_DG_plausibility}.}
    \label{tab:appendix_DG_ablation_study}
\end{table}

\subsection{Ablation Study} \label{sec:appendix_DG_ablation_study}
The training settings used for the ablation study are identical to those of our distractor generator training setup (Appendix~\ref{sec:appendix_DG_settings}), except that the base MCQ dataset was used as the training data instead of the student choice dataset.
Table~\ref{tab:appendix_DG_ablation_study} presents the results of the ablation study (compare with Table~\ref{tab:evaluation_DG_plausibility} and~\ref{tab:evaluation_DG_plausibility_per_question}).

\subsection{Error Analysis} \label{sec:appendix_DG_error_analysis}
Analyzing the low-quality samples generated by our distractor generator revealed the following types of errors:

First, the model sometimes failed to produce the specified number of distractors based on the input parameter $n$, or it created duplicate distractors among the outputs.

Next, for code type questions, the generated distractors lacked diversity in output formats and often made minimal changes, such as altering only one or two variables, resulting in repetitive and insufficiently varied distractors.

Meanwhile, for statement type questions, the model overly mimicked the correct answer, creating distractors based on only one or two concepts, while failing to effectively incorporate other related concepts.

Future work to improve the distractor generator could involve explicitly providing the model with information on \textit{similar concepts} or \textit{common errors} that students are likely to confuse.


\subsection{Prompt for Checking Distractor Validity} \label{sec:appendix_DG_validity}
The instruction prompt for checking the validity of distractors is in Table~\ref{tab:appendix_DG_validity}.
If the output is `invalid' (as it is an incorrect option for the question), it is considered a distractor.

\section{Base MCQ Dataset}
We were provided with an MCQ dataset by an online learning platform for educational research purposes and processed it for use within the scope of the provided purpose. The questions and options, originally in Korean, were translated into English for experimental purposes. The provided MCQ data does not contain any personally identifiable information about the individuals who answered the questions, and we manually checked to confirm that the text does not include any offensive content.


\section{Prompt for Augmenting Distractors in the Base MCQ Dataset} \label{sec:appendix_SCD_augment}
The instruction prompt for augmenting distractors in the base MCQ dataset is in Table~\ref{tab:appendix_SCD_augment}. Through this prompt, the student choice dataset was constructed only when at least one newly generated distractor by GPT-4o was valid and did not overlap with the original.

\section{Potential Issues}
MCQs serve as a tool for assessing students' knowledge, so the options must be based on accurate information (i.e., both the correct answer and distractors must be valid). As mentioned earlier in the limitations, distractors generated by the model may not be actual incorrect options to the question. To proactively address the potential issue, we explored methodologies to ensure the validity of the distractors generated by the model. As part of these efforts, we implemented instruction prompts and output formats for the model to classify the type ($T$) of distractors, thereby mitigating this issue.

We used selection rate data from questions answered by hundreds of students to ensure the reliability of common misconception information for training the pairwise ranker. However, since misconceptions can vary by learning level or educational environment, the model's reasoning may not generalize to other populations. To make accurate predictions for a target population, selection rates specific to that group should be used.

\begin{table*}[t!]
    \centering
    \small
    \renewcommand{\arraystretch}{1.5} 
    \begin{tabular}{p{0.95\textwidth}}
        \Xhline{2\arrayrulewidth}
        \textbf{Pairwise Ranker Prompt (Reasoning)} \\ 
        \hline
        [INST] You are a teacher analyzing which distractor in a given Multiple Choice Question is more confusing for students and why. Your review should include the following content in one paragraph: \newline
        - Describe a realistic process of solving the problem from a student's perspective as you look at each distractor.\newline
        - Consider why it might be plausible as the correct/incorrect statement, based on students' misconceptions, mistakes, intuition, etc., from various angles.\newline
        Output your choice as a single token, either A or B, that students are more likely to choose.\newline\newline
        [Question] \{\textit{question}\}\newline
        [Answer] \{\textit{answer}\}\newline
        [Distractor A] \{\textit{distractor}\}\newline
        [Distractor B] \{\textit{distractor}\}\newline\newline
        Generate in the following format:\newline
        \#\#\# Review:\newline
        \#\#\# Choice: [/INST]
        \\
        \Xhline{2\arrayrulewidth}
    \end{tabular}
    \caption{Instruction prompt (Reasoning) for pairwise ranker.}
    \label{tab:appendix_PR_prompt_reasoning}
\end{table*}

\begin{table*}[t!]
    \centering
    \small
    \renewcommand{\arraystretch}{1.5} 
    \begin{tabular}{p{0.95\textwidth}}
        \Xhline{2\arrayrulewidth}
        \textbf{Pairwise Ranker Prompt (Rubric)} \\ 
        \hline
        Analyze which side of the given Multiple Choice Question distractor pair is more confusing and plausible to students based on the given rubric.\newline\newline
        [Question] \{\textit{question}\}\newline
        [Answer] \{\textit{answer}\}\newline
        [Distractor A] \{\textit{distractor}\}\newline
        [Distractor B] \{\textit{distractor}\}\newline\newline
        Evaluation Rubric:\newline
        [1]. Conceptual Misunderstandings: Evaluate if the distractor addresses into specific misconceptions or partial understandings related to the question.\newline
        [2]. Similarity to Correct Answer: Assess how closely the distractor resembles the correct answer, either in structure, terminology, or context.\newline
        [3]. Intuitive Appeal: Analyze if the distractor seems logical or intuitively correct based on common language use or student intuition.\newline\newline
        Generation Guide:\newline
        - [n]: For each evaluation criterion, review in one sentence how each distractor may or may not confuse students.\newline
        - [Summary]: Summarize the review, and choose more confusing and plausible distractor.\newline
        - [Choice]: Output your choice as a single token, either A or B.\newline\newline
        Generate in the following format:\newline
        [1]:\newline
        [2]:\newline
        [3]:\newline
        [Summary]:\newline
        [Choice]:
        \\ 
        \Xhline{2\arrayrulewidth}
    \end{tabular}
    \caption{Instruction prompt (Rubric) for pairwise ranker.}
    \label{tab:appendix_PR_prompt_rubric}
\end{table*}

\begin{table*}[t!]
    \centering
    \small
    \renewcommand{\arraystretch}{1.5} 
    \begin{tabular}{p{0.95\textwidth}}
        \Xhline{2\arrayrulewidth}
        \textbf{Pairwise Ranker Prompt (G-Eval)} \\ 
        \hline
        You will be given one multiple-choice question (MCQ) and two distractors. Your task is to choose one distractor based on the metric.\newline
        Please make sure you read and understand these instructions carefully. Please keep this document open while reviewing, and refer to it as needed.\newline\newline
        Evaluation Criteria:\newline
        Plausibility: This metric indicates how likely students are to feel that the distractor is the correct answer and choose it. A distractor with high plausibility is similar in form to the correct answer or contains common misconceptions and mistakes, making students more likely to select it.\newline\newline
        Evaluation Steps:\newline
        1. Read the MCQ carefully and think about the relevant misconceptions or mistakes related to the question from your perspective as a teacher.\newline
        2. Judge how plausible and confusing the distractor would be from a student's perspective.\newline
        3. Choose one distractor based on Evaluation Criteria. Output your choice as a single token, either A or B.\newline\newline
        [Question] \{\textit{question}\}\newline
        [Answer] \{\textit{answer}\}\newline
        [Distractor A] \{\textit{distractor}\}\newline
        [Distractor B] \{\textit{distractor}\}\newline\newline
        Evaluation Form (A or B ONLY):\newline
        - Choice:
        \\ 
        \Xhline{2\arrayrulewidth}
    \end{tabular}
    \caption{Instruction prompt (G-Eval) for pairwise ranker.}
    \label{tab:appendix_PR_prompt_g-eval}
\end{table*}

\begin{table*}[t!]
    \centering
    \small
    \renewcommand{\arraystretch}{1.5} 
    \begin{tabular}{p{0.95\textwidth}}
        \Xhline{2\arrayrulewidth}
        \textbf{Pairwise Ranker Prompt (Discussion)} \\ 
        \hline
        \textbf{<Prompt - Student>}\newline
        Play the role of students with three different levels of proficiency: A is low, B is medium, and C is high.\newline
        A lower proficiency level indicates more confusion about the concept, while a higher proficiency level indicates a better understanding of the related knowledge.\newline
        - In a cooperative learning situation, three students with different levels of proficiency are discussing and solving a given problem together.\newline
        - For each option in the MCQ, share your thoughts according to each proficiency level. Discuss similar concepts and any confused or mistaken knowledge, ask for help, give advice, and interact actively.\newline
        - Having a high proficiency level does not mean knowing the correct answer. However, they have better problem-solving skills through reasonable inference.\newline
        - Take turns speaking equally among the low, medium, and high proficiency students. Use natural transitions like 'Shall we talk about this option next?' to keep the discussion flowing smoothly. End the discussion after discussing all the options.\newline\newline
        [Question] \{\textit{question}\}\newline
        [Options] \{\textit{distractors}\}\newline\newline
        Output the result in the following format:\newline
        [A]: "..."\newline
        [B]: "..."\newline
        [C]: "..." \newline\newline
        \textbf{<Prompt - Teacher>}\newline
        Act as teachers discussing and judging the plausibility (whether it would confuse students) score of each distractor in a given MCQ.\newline 
        - First, analyze the collaborative learning records of three students. Then, as the first teacher, choose between distractors A and B, deciding which one is more likely to confuse students or be frequently selected by them.\newline
        - As the second teacher, share your thoughts and provide reasonable counterarguments. Use the collaborative learning records of the three students as supporting evidence for your scoring. The second teacher should always question the initial score, challenge generalized assumptions, and argue which distractor is more plausible.\newline
        - Take turns discussing and adjusting the choice.\newline
        - The utterances must be clear and concise.\newline\newline
        [Question] \{\textit{question}\}\newline
        [Answer] \{\textit{answer}\}\newline
        [Cooperative Learning Records] \{\textit{cooperative learning records}\}\newline
        [Distractor A] \{\textit{distractor}\}\newline
        [Distractor B] \{\textit{distractor}\}\newline
        [Discussion History] \{\textit{history}\}\newline\newline
        A conclusion must be reached within a maximum of 5 utterances, taking into account both [T1] and [T2] combined.\newline
        Once you both agree on the final choice, output \#\#\# Choice: A or \#\#\# Choice: B.\newline
        Generate the next utterance in the discussion based on the discussion history:\newline
        [T1 or T2]: "..."
        \\ 
        \Xhline{2\arrayrulewidth}
    \end{tabular}
    \caption{Instruction prompt (Discussion) for pairwise ranker.}
    \label{tab:appendix_PR_prompt_discussion}
\end{table*}

\begin{table*}[t!]
    \centering
    \small
    \renewcommand{\arraystretch}{1.5} 
    \begin{tabular}{p{0.95\textwidth}}
        \Xhline{2\arrayrulewidth}
        \textbf{Pairwise Ranker Prompt \citep{scarlatos-etal-2024-improving}} \\ 
        \hline
        [INST] A teacher assigns the following programming question to the students.\newline
        Question: \{\textit{question}\}\newline
        Correct answer: \{\textit{answer}\}\newline
        Generate a distractor for this question that targets some student misconception.\newline
        Distractor: [/INST] \{\textit{distractor}\}
        \\ 
        \Xhline{2\arrayrulewidth}
    \end{tabular}
    \caption{Instruction prompt \citep{scarlatos-etal-2024-improving} for pairwise ranker.}
    \label{tab:appendix_PR_prompt_scarlatos}
\end{table*}

\begin{table*}[t!]
    \centering
    \small
    \renewcommand{\arraystretch}{1.5} 
    \begin{tabular}{p{0.95\textwidth}}
        \Xhline{2\arrayrulewidth}
        \textbf{Prompt for Generating Pairwise Ranker Training Data} \\ 
        \hline
        You are a teacher analyzing which distractor in a given Multiple Choice Question is more confusing for students and why.\newline
        Your review should include the following content in one paragraph:\newline
        - Describe a realistic process of solving the problem from a student's perspective as you look at each distractor. Consider why it might be plausible as the correct/incorrect statement, based on students' misconceptions, mistakes, intuition, etc., from various angles.\newline
        - Output your choice as a single token, either A or B, that students are more likely to choose.\newline\newline
        [Question] \{\textit{question}\}\newline
        [Answer] \{\textit{answer}\}\newline
        [Distractor A] \{\textit{distractor a}\}\newline
        [Distractor B] \{\textit{distractor b}\}\newline
        Distractor chosen more frequently by actual students:\{\textit{a or b}\}\newline\newline
        Make sure your choice matches the distractor most frequently chosen by actual students. However, you must not mention this information as if you originally knew it.\newline
        Generate in the following format:\newline
        \#\#\# Review: \newline 
        \#\#\# Choice: \\ 
        \Xhline{2\arrayrulewidth}
    \end{tabular}
    \caption{Instruction prompt for generating pairwise ranker training data.}
    \label{tab:appendix_PR_training}
\end{table*}

\begin{table*}[t!]
    \centering
    \small
    \renewcommand{\arraystretch}{1.5} 
    \begin{tabular}{>{\raggedright\arraybackslash}p{0.18\textwidth}|p{0.72\textwidth}}
        \Xhline{2\arrayrulewidth}
        \textbf{Category} & \textbf{Definition} \\ 
        \hline
        Operation Confusion 
        &
        Distractors that involve misunderstanding of specific operations, such as incorrect assumptions about function outputs or operation precedence. 
        \\ \hline
        Structure Error
        &
        Distractors reflecting improper syntax or structural misunderstandings. 
        \\ \hline
        Calculation Mistake
        &
        Distractors that exploit errors in arithmetic, index calculations, or logical evaluations, leading to incorrect results. 
        \\ \hline
        Syntax Familiarity
        &
        Distractors that align with common syntax conventions or structures from Python or other programming languages, leading to confusion due to familiarity. 
        \\ \hline
        Logical Consistency
        &
        Distractors that maintain a consistent or plausible logic or pattern, even if incorrect, which can mislead students who are not fully confident in their understanding. 
        \\ \hline
        Additional Conditions
        &
        Distractors that introduce extra conditions or columns, which may lead students to misinterpret the problem as requiring more complex logic, thus creating confusion. 
        \\
        \Xhline{2\arrayrulewidth}
    \end{tabular}
    \caption{Definitions of plausibility factors of code type question.}
    \label{tab:appendix_PR_factors_code}
\end{table*}
\begin{table*}[t!]
    \centering
    \small
    \renewcommand{\arraystretch}{1.5} 
    \begin{tabular}{>{\raggedright\arraybackslash}p{0.18\textwidth}|p{0.72\textwidth}}
        \Xhline{2\arrayrulewidth}
        \textbf{Category} & \textbf{Definition} \\ 
        \hline
        Ambiguity and Complexity
        &
        Distractors that introduce nuanced or ambiguous details, leading to confusion and misinterpretation due to their complexity or lack of clarity. 
        \\ \hline
        Conceptual Overlap
        &
        Distractors that involve concepts or operations that overlap with other similar terms, causing students to conflate them and mistakenly believe they are correct. 
        \\ \hline
        Familiarity Traps
        &
        Distractors that use familiar terms or straightforward statements, making them seem correct at first glance and less likely to be critically analyzed by students. 
        \\ \hline
        Partial Understanding
        &
        Distractors built on incomplete knowledge, leading students to make errors due to gaps in conceptual clarity. 
        \\ \hline
        Overgeneralization
        &
        Distractors that appear plausible by relying on students' tendency to apply learned concepts too broadly without verifying their validity in specific contexts.  
        \\ \hline
        Practical Experience
        &
        Distractors that leverage students' familiarity with common tasks, such as data manipulation or querying, creating false confidence in their correctness.  
        \\
        \Xhline{2\arrayrulewidth}
    \end{tabular}
    \caption{Definitions of plausibility factors of statement type question.}
    \label{tab:appendix_PR_factors_statement}
\end{table*}

\begin{table*}[t!]
    \centering
    \small
    \renewcommand{\arraystretch}{1.5} 
    \begin{tabular}{p{0.95\textwidth}}
        \Xhline{2\arrayrulewidth}
        \textbf{Human Evaluation Survey Form} \\ 
        \hline
        \textbf{<Guideline>}\newline
        The following provides a programming multiple-choice question, along with an analysis (review) that predicts which of the two incorrect options is more challenging for students (i.e., more likely to be chosen).
        You are tasked with evaluating the quality of the analysis from the perspective of an education expert and stating whether you agree with the analysis.\newline\newline
        Provided Items:\newline
        [Question]: The question\newline
        [Answer]: The correct answer\newline
        [Distractor A and B]: The two incorrect options, A and B\newline
        [Review]: An analysis of which incorrect option (A or B) would be more confusing (more likely to be chosen) by students, along with the final selection\newline\newline
        Evaluation Criteria:\newline
        - Logical Reasoning: Whether the reasoning process is logical.\newline
        - Student Understanding: Whether the reasoning effectively understands students' misconceptions or problem-solving processes.\newline
        - Knowledge Accuracy: Whether the reasoning is based on accurate and error-free knowledge.\newline
        - Choice Agreement: Whether the evaluator agrees with the model's final choice.\newline\newline
        \textbf{<Item>}\newline
        [Question] \{\textit{question}\}\newline
        [Answer] \{\textit{answer}\}\newline
        [Distractor A] \{\textit{distractor}\}\newline
        [Distractor B] \{\textit{distractor}\}\newline
        [Review] \{\textit{model's reasoning}\}\newline\newline
        - The reasoning process in the review is logical.\newline 
        | 1. Strongly Disagree | 2. Disagree | 3. Neutral | 4. Agree | 5. Strongly Agree |\newline
        - The review demonstrates a good understanding of actual student misconceptions or problem-solving processes.\newline
        | 1. Strongly Disagree | 2. Disagree | 3. Neutral | 4. Agree | 5. Strongly Agree |\newline
        - The review is based on accurate and error-free knowledge.\newline
        | 1. Strongly Disagree | 2. Disagree | 3. Neutral | 4. Agree | 5. Strongly Agree |\newline
        - I agree with the final choice in the review.\newline
        | 1. Strongly Disagree | 2. Disagree | 3. Neutral | 4. Agree | 5. Strongly Agree | \\ 
        \Xhline{2\arrayrulewidth}
    \end{tabular}
    \caption{Survey form for human evaluation on the pairwise ranker. The original guideline in Korean has been translated into English.}
    \label{tab:appendix_PR_human_evaluation}
\end{table*}

\begin{table*}[t!]
    \centering
    \small
    \renewcommand{\arraystretch}{1.5} 
    \begin{tabular}{p{0.95\textwidth}}
        \Xhline{2\arrayrulewidth}
        \textbf{Pairwise Ranker Prompt (Ablation Study, w/o Reasoning)} \\ 
        \hline
        [INST] You are a teacher analyzing which distractor in a given Multiple Choice Question is more confusing for students.
        Output your choice as a single token, either A or B, that students are more likely to choose.\newline\newline
        [Question] \{\textit{question}\}\newline
        [Answer] \{\textit{answer}\}\newline
        [Distractor A] \{\textit{distractor}\}\newline
        [Distractor B] \{\textit{distractor}\}\newline\newline
        Generate in the following format:\newline
        \#\#\# Choice: [/INST] \\ 
        \Xhline{2\arrayrulewidth}
    \end{tabular}
    \caption{Instruction prompt for ablation study on the pairwise ranker.}
    \label{tab:appendix_PR_ablation_study}
\end{table*}

\begin{table*}[t!]
    \centering
    \small
    \renewcommand{\arraystretch}{1.5} 
    \begin{tabular}{p{0.95\textwidth}}
        \Xhline{2\arrayrulewidth}
        \textbf{Distractor Generator Prompt (Ours)} \\ 
        \hline
        [INST] You are a teacher tasked with creating distractors (plausible wrong options) for a given Multiple Choice Question.\newline
        Generate distractors according to the guide below:\newline
        1) Distractor type:\newline
        - Analyze whether the question asks for a `correct' or `incorrect' option.\newline
        - If the question asks for a correct option, the distractor type should be "Incorrect knowledge"; if it asks for an incorrect option, the distractor type should be "Correct knowledge".\newline
        2) Distractors:\newline
        - The distractor should be well-formatted so that it fits naturally when presented together with the question and answer.\newline
        - If the distractor type is "Incorrect knowledge", the distractor must be an actually incorrect statement; if the distractor type is "Correct knowledge", the distractor must be an actually correct statement.\newline\newline
        [Question] \{\textit{question}\}\newline
        [Answer] \{\textit{answer}\}\newline\newline
        Generate \{\textit{n}\} distractor(s) in the following format:\newline
        \#\#\# Type: \newline
        \#\#\# Distractor n: [/INST]
        \\ 
        \Xhline{2\arrayrulewidth}
    \end{tabular}
    \caption{Instruction prompt for distractor generator (Ours).}
    \label{tab:appendix_DG_prompt}
\end{table*}

\begin{table*}[t!]
    \centering
    \small
    \renewcommand{\arraystretch}{1.5} 
    \begin{tabular}{p{0.95\textwidth}}
        \Xhline{2\arrayrulewidth}
        \textbf{Distractor Generator Prompt (kNN approach by \citet{feng-2024-exploring})} \\ 
        \hline
        Question: \{\textit{in-context question}\}\newline
        Answer: \{\textit{in-context answer}\}\newline
        Distractor1: \{\textit{in-context distractor}\}\newline
        Distractor2: \{\textit{in-context distractor}\}\newline
        Distractor3: \{\textit{in-context distractor}\}\newline
        \newline
        Question: \{\textit{in-context question}\}\newline
        Answer: \{\textit{in-context answer}\}\newline
        Distractor1: \{\textit{in-context distractor}\}\newline
        Distractor2: \{\textit{in-context distractor}\}\newline
        Distractor3: \{\textit{in-context distractor}\}\newline
        \newline
        Question: \{\textit{in-context question}\}\newline
        Answer: \{\textit{in-context answer}\}\newline
        Distractor1: \{\textit{in-context distractor}\}\newline
        Distractor2: \{\textit{in-context distractor}\}\newline
        Distractor3: \{\textit{in-context distractor}\}\newline
        \newline
        Referencing the above samples, generate 3 distractors.\newline
        Question: \{\textit{question}\}\newline
        Answer: \{\textit{answer}\}\newline
        Distractor1:\newline
        Distractor2:\newline
        Distractor3:
        \\ 
        \Xhline{2\arrayrulewidth}
    \end{tabular}
    \caption{Instruction prompt for distractor generator (kNN approach).}
    \label{tab:appendix_DG_knn_prompt}
\end{table*}

\begin{table*}[t!]
    \centering
    \small
    \renewcommand{\arraystretch}{1.5} 
    \begin{tabular}{p{0.95\textwidth}}
        \Xhline{2\arrayrulewidth}
        \textbf{Prompt for Checking Distractor Validity} \\ 
        \hline
        Check if the given option is the correct choice in a multiple-choice question (MCQ).\newline
        1. Check whether the question asks for a `correct' or `incorrect' option. If the question asks for a correct option, label "type" as "asking correct option." If the question asks for an incorrect option, label "type" as "asking incorrect option."\newline
        2. Insert the given option into the question and analyze whether it is the correct choice.\newline
        3. Based on the analysis, if the option is the correct answer to the question, label it as "valid." If it is not the correct answer, label it as "invalid."\newline\newline
        [Question] \{\textit{question}\}\newline
        [Option] \{\textit{distractor}\}\newline\newline
        Output according to the following JSON format:\newline
        \{\{\newline  
            "type": "asking correct option" or "asking incorrect option",\newline
            "analysis": "your analysis in one sentence",\newline
            "validity": "valid" or "invalid"\newline  
        \}\} \\ 
        \Xhline{2\arrayrulewidth}
    \end{tabular}
    \caption{Instruction prompt for checking the validity of distractors.}
    \label{tab:appendix_DG_validity}
\end{table*}

\begin{table*}[t!]
    \centering
    \small
    \renewcommand{\arraystretch}{1.5} 
    \begin{tabular}{p{0.95\textwidth}}
        \Xhline{2\arrayrulewidth}
        \textbf{Prompt for Augmenting Distractors in the Base MCQ Dataset} \\ 
        \hline
        You are a teacher tasked with creating distractors (plausible wrong options) for a given Multiple Choice Question.\newline
        Generate distractors according to the guide below:\newline
        1) Distractor type:\newline
        - Analyze whether the question asks for a `correct' or `incorrect' option.\newline
        - If the question asks for a correct option, the distractor type should be "Incorrect knowledge"; if it asks for an incorrect option, the distractor type should be "Correct knowledge".\newline
        2) Distractors:\newline
        - The distractor should be well-formatted so that it fits naturally when presented together with the question and answer.\newline
        - If the distractor type is "Incorrect knowledge", the distractor must be an actually incorrect statement; if the distractor type is "Correct knowledge", the distractor must be an actually correct statement.\newline
        - Refer to the original distractors provided.\newline\newline
        [Question] \{\textit{question}\}\newline
        [Answer] \{\textit{answer}\}\newline
        [Original Distractors] \{\textit{distractors}\}\newline\newline
        Generate 3 new distractor(s) in the following JSON format:\newline
        \{\{\newline 
            "type": "Incorrect knowledge" or "Correct knowledge",\newline
            "distractor\_n": "n-th distractor in string type",\newline
            ...\newline  
        \}\} \\ 
        \Xhline{2\arrayrulewidth}
    \end{tabular}
    \caption{Instruction prompt for augmenting distractors in the base MCQ dataset.}
    \label{tab:appendix_SCD_augment}
\end{table*}

\end{document}